\documentclass[letterpaper, 10 pt, journal, twoside]{ieeetran} 
\usepackage{amsmath,amssymb,amsfonts}
\usepackage{algorithmic}
\usepackage{graphicx}
\usepackage{textcomp}
\usepackage{color}
\usepackage{subfigure}
\usepackage{url}
\usepackage[ruled,linesnumbered]{algorithm2e}
\usepackage{array}
\usepackage{multirow}
\usepackage{booktabs}
\usepackage{makecell}
\usepackage{threeparttable}

\begin{document}

\markboth{IEEE Robotics and Automation Letters. Preprint Version. Accepted August, 2025}
{Chen \MakeLowercase{\textit{et al.}}: FDSPC: Fast and Direct Smooth Motion Planning}

\title{FDSPC: Fast and Direct Smooth Motion Planning via Continuous Curvature Integration}

\author{Zong Chen$^{1}$, Haoluo Shao$^{1}$, Ben Liu$^{1}$, Siyuan Qiao$^{1}$, Yu Zhou$^{2}$, and Yiqun Li$^{1}$%
  \thanks{Manuscript received: March, 31, 2025;  Revised July, 7, 2025; Accepted August, 4, 2025.}
  \thanks{This paper was recommended for publication by Editor Aniket Bera upon evaluation of the Associate Editor and Reviewers' comments. This work was supported by the National Natural Science Foundation of China No. 51905185, the Interdisciplinary Research Support Program of HUST No. 2024JCYJ040, and the National Postdoctoral Program for Innovative Talents No. BX20180109. \textit{(Corresponding author: Yiqun Li.)}}%
  \thanks{$^{1}$Zong Chen, Haoluo Shao, Siyuan Qiao and Yiqun Li are with the State Key Laboratory of Intelligent Manufacturing Equipment and Technology, School of Mechanical Science and Engineering, Huazhong University of Science and Technology, Wuhan, 430074, China. (e-mail:{\tt\small \{skelon\_chan; shao\_haoluo; m202470713; siyuan\_q; liyiqun\}@hust.edu.cn})}%
  \thanks{$^{2}$Yu Zhou is with School of Electronic Information and Communications, HUST 430074, China (e-mail: \tt\small yuzhou@hust.edu.cn)}%
}

\maketitle

\begin{abstract}
In recent decades, mobile robot motion planning has seen significant advancements. Both search-based and sampling-based methods have demonstrated capabilities to find feasible solutions in complex scenarios. Mainstream path planning algorithms divide the map into occupied and free spaces, considering only planar movement and ignoring the ability of mobile robots to traverse obstacles in the $z$-direction. Additionally, paths generated often have numerous bends, requiring additional smoothing post-processing. In this work, a fast, and direct motion planning method based on continuous curvature integration that takes into account the robot's obstacle-crossing ability under different parameter settings is proposed. This method generates smooth paths directly with pseudo-constant velocity and limited curvature, and performs curvature-based speed planning in complex $2.5$-D terrain-based environment (take into account the ups and downs of the terrain), eliminating the subsequent path smoothing process and enabling the robot to track the path generated directly. The proposed method is also compared with some existing approaches in terms of solution time, path length, memory usage and smoothness under multiple scenarios. The proposed method is vastly superior to the average performance of state-of-the-art (SOTA) methods, especially in terms of the self-defined $\mathcal{S}_2 $ smoothness (mean angle of steering). Furthermore, simulations and experiments are conducted on our self-designed wheel-legged robot with $2.5$-D traversability. These results demonstrate the effectiveness and superiority of the proposed approach in several representative environments.\footnote{The implementation of this work is available at \url{https://github.com/SkelonChan/GPCC_curvature_planning.}}
\end{abstract}


\IEEEpeerreviewmaketitle

\section{Introduction}
\IEEEPARstart{R}{obot} motion planning has undergone significant development in recent years, and played crucial roles in various fields, such as autonomous vehicles, robot arms and unmanned aerial vehicles. However, existing commonly used path planning methods, such as search-based algorithms A* \cite{hart1968formal}, Dijkstra \cite{dijkstra2022note}, sampling-based algorithms Rapidly-exploring Random Trees (RRT), RRT*, extended-RRT, RRT-Connect \cite{karaman2011anytime} and swarm intelligence-based algorithms Ant Colony Optimization (ACO) \cite{chen2021path}, all yield non-continuous, zigzag global paths. Except for end-to-end approaches for robot navigation \cite{han2024neupan, frossard2018end}, the majority of these methods require additional smoothing or post-processing to be effectively applied to robot trajectory tracking. Moreover, during the smoothing or optimization process, the newly generated path will inevitably deviate from the original collision-free path in certain regions, resulting in additional collision re-detection and replanning \cite{li2023trajectory}. Gradient-based methods \cite{ratliff2009chomp}, convex optimization \cite{marcucci2023motion}, and SQP-based approaches \cite{schulman2013finding} can generate smooth trajectories, but often struggle in complex environments due to high-dimensional gradient computations, reliance on convex decomposition, and sensitivity to non-convexity that demands costly iterative solving.

\begin{figure}[t]
  \centering
  \subfigure{
    \includegraphics[width=0.98\linewidth]{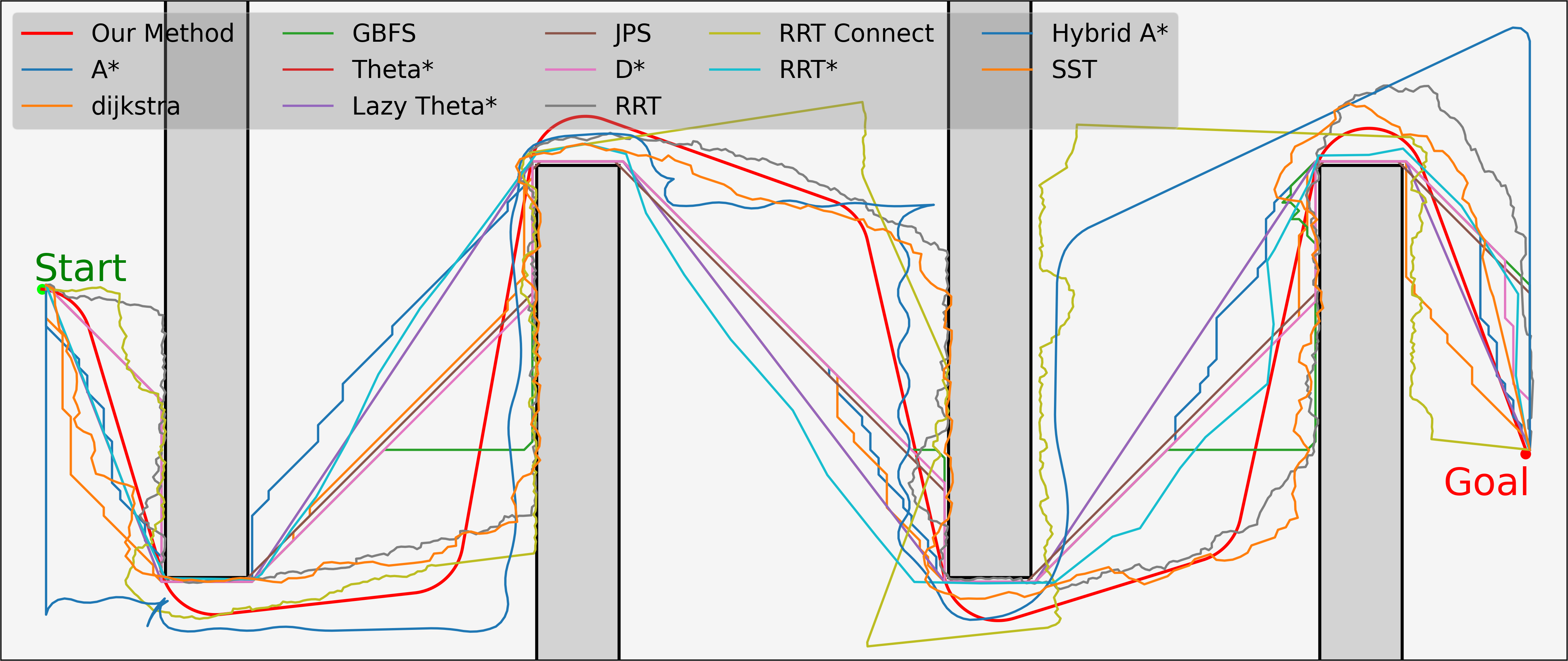}
  }
  \hspace{-5.5mm} 
  \subfigure{
    \includegraphics[width=0.98\linewidth]{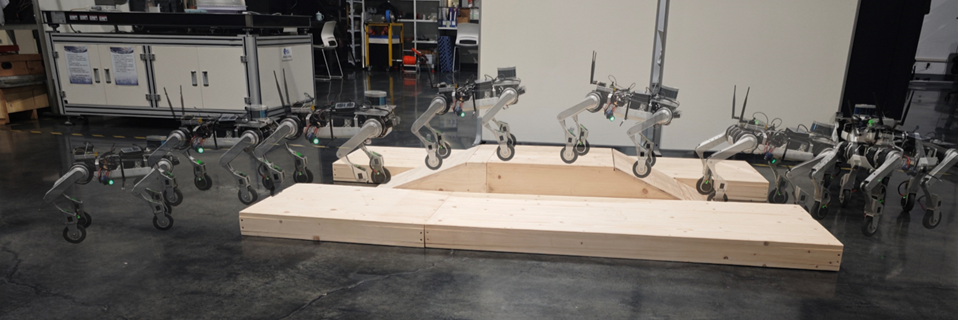}
  }
  \caption{Comparison of the planning results of the proposed FDSPC algorithm with other state-of-the-art path planning algorithms in simple maze (Up) and the experiment in $2.5$-D terrain-based environment by a wheel-legged robot with four independent steering wheels (Bottom).}
  \label{fig1}
\end{figure}
\par
In this letter, a fast and direct motion planning method based on continuous curvature integration (FDSPC) is proposed for mobile robot trajectory tracking on a given map, as shown in Fig.~\ref{fig1}. The algorithm iteratively explores collision-free path segments satisfying $G^2$ smoothness \cite{farin2001curves} (curvature continuity). If trapped in a local solution, it automatically backtracks to the previous optimal state to ensure feasibility. FDSPC demonstrates superior performances by comparing various indicators, including solution time, smoothness, in multiple scenarios. The contributions of this letter are as follows,
\begin{enumerate}
  \item A fast motion planning method based on continuous curvature integration is proposed, which can generate global paths that satisfies $G^2$ smoothness, avoiding the re-collision checking and smoothing.
  \item A variant of the direct positioning binary tree combined with an ordered dictionary is introduced to facilitate heuristic search of the path rapidly and ensuring both the feasibility and efficiency of the algorithm.
  \item A variety of evaluation indicators are compared across multiple scenarios, and successfully applied in our self-designed wheel-legged robot with four independent steering wheels in $2.5$-D terrain-based  environments.
\end{enumerate}

\section{Related works}
\subsection{Classical motion planning}
The motion planning problem has been around for decades, and the methods can be broadly divided into two categories, i.e., the search-based approaches and the sampling-based approaches. The former approaches discretize the environment and searches for the optimal path in the grid map. The latter methods sample in the configuration space and then find a feasible path through the sampling points. Besides, the paths generated by swarm intelligence optimization \cite{shi2022path} are more prone to fall into local optimum, and the optimization-based methods \cite{zuo2020mpc} minimize a predefined cost function under various constraints but require a good initial value; otherwise, their efficiency and feasibility are difficult to guarantee.
\par
The Greedy Best-First Search (GBFS) and $\theta^*$ family methods, including Lazy $\theta^*$, derive from classic A*-based grid search and are well described in Planning Algorithms \cite{uras2015empirical}, where GBFS prioritizes heuristic cost leading to fast but possibly suboptimal paths, $\theta^*$ improves smoothness by allowing connections between non-adjacent nodes, and Lazy $\theta^*$ further enhances efficiency by delaying visibility checks. The $D^*$ algorithm, also covered in \cite{stentz1994optimal}, maintains a priority queue to enable efficient path replanning in dynamic environments. The Stable Sparse Rapidly-exploring Random Tree (SST) \cite{li2016asymptotically} uses sparse sampling to reduce storage and computational complexity while ensuring asymptotic optimality.
\subsection{Smooth path planning}
Building upon the preceding path results, the curve planning method is crafted by amalgamating pathfinding and smoothness through the incorporation of Reed-Sheep (RS) curves \cite{reeds1990optimal}, B\'{e}zier curves, or similar techniques. The hybrid A* \cite{sedighi2019guided} smooths the A*-derived path using RS curves, ensuring that the kinematic constraints of the vehicle are followed and achieving $G^1$ continuity (tangency continuity). \cite{zhang2021hierarchical, zhang2019hybrid} propose a hierarchical search spatial scales-based hybrid A* (HHA*) framework, which employs clothoids instead of RS curves, attaining $G^2$ continuity of the path and exhibiting commendable performance in parking scenario. Clothoids with continuous curvature expand the original motion primitive and enhance the granularity of the search space. However, clothoids are defined in terms of Fresnel integrals \cite{abramowitz1965handbook}, making them challenging to be applied online. To address this, a lookup table is constructed to store the coordinates of basic clothoids \cite{brezak2013real}, which accelerates calculations when dealing with other clothoids, yielding lower errors compared to approximations using B\'{e}zier or B-spline curves \cite{wang2001approximation}. Nevertheless, achieving an exact approximation for the clothoid is not crucial in path planning, especially in iterative approaches, and none of the aforementioned algorithms can generate a smooth, and applicable trajectory directly and rapidly.
\par
In this letter, a fast and direct motion planning method FDSPC, which can generate $G^2$ continuity trajectory without extra smoothing and take into account the obstacle-crossing ability of different robots is proposed. Unlike classical methods that plan directly in configuration or state space, FDSPC is built upon curvature-based and planning in the configuration space accordingly.
\section{Curvature planning}
The representation of a path in $n$-dimensional ($n$-D) space can be expressed as the set $\mathcal{C}$ of coordinates of discrete points $p\in\mathbb{R}^n$. For any curve in $n$-D space, it can be described by an $(n-1)$-D function, precisely. For example, on a $2$-D plane, the curve can be described by a $1$-D curvature $\kappa$, while in $3$-D space, the curve satisfies the continuity of curvature and torsion \cite{guggenheimer2012differential}, and be twice differentiable in all directions by solving the $Frenet-Serret$ equations. In $n$ dimensional space, this requires solving $(n-1)$ coupled differential equations and integrating the tangent vector to obtain the path, making the process much more complex than $2$ or $2.5$-D spaces. Therefore, in this section, we focus solely on curvature planning in $2$-D plane and $2.5$-D terrain space.
\par
\subsection{Curvature planning on $2$-D plane}
The relative positional relationship between path points on $2$-D plane and the obstacles is not directly discernible through the curvature. Thus, it is necessary to unfold the implicit expression, i.e., map the curvature function $\kappa$ to the path points by double integrating $\kappa$ on $2$-D plane, then evaluate the feasibility (whether the path generated collides with obstacles) and adjust the curvature accordingly. The relationship between curvature and the tangent at the path points can be expressed as,
\begin{equation}
  \label{eq1}
  \frac{d\theta}{ds}=\kappa \\\Rightarrow \frac{d\theta}{dt}\frac{dt}{ds}=\kappa \\\Rightarrow \frac{d\theta}{dt}=\kappa \frac{ds}{dt}
\end{equation}
where $\theta$ is the tangent angle, $s$ is the arc length of the path, $dt$ is the integration step, $\kappa$ is the path curvature, and ${ds}/{dt}$ can be regarded as the pseudo velocity of the path (the rate of change of distance along the curve \cite{tu2017differential}). The tangent angle $\theta(t)$ can be obtained by integrating the curvature function $\kappa(t)$ and the pseudo velocity ${ds}/{dt}$.
\par
\begin{figure}[t]
  \centering
  \subfigure{
    \includegraphics[width=0.48\linewidth]{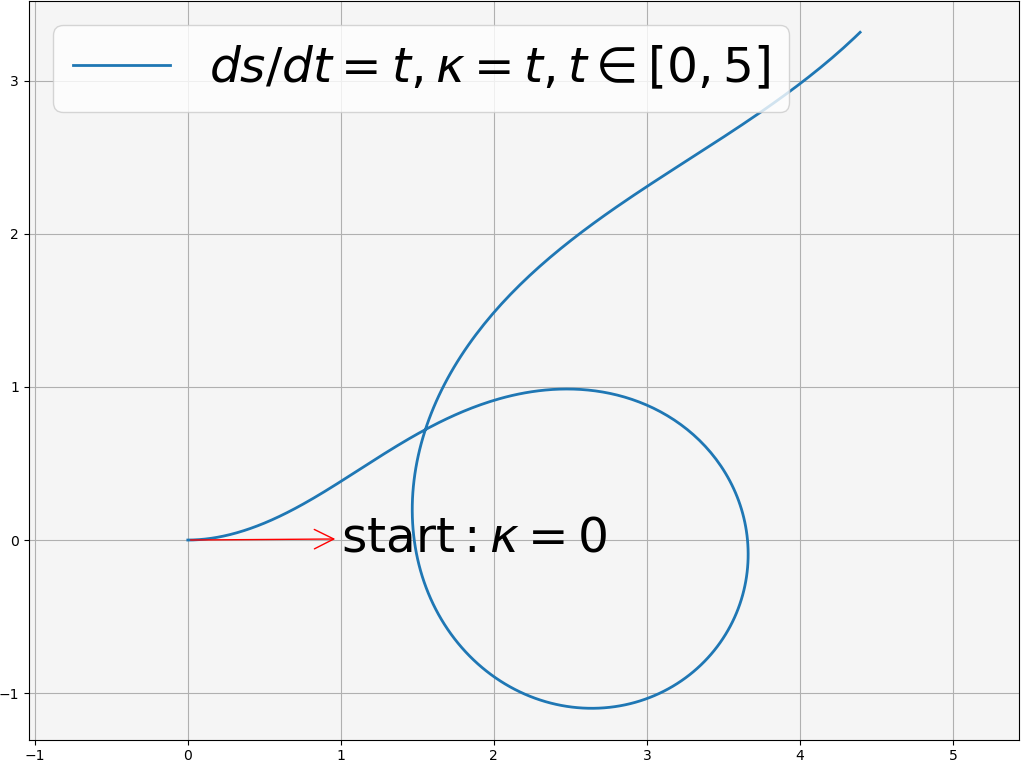}
  }
  \hspace{-4mm} 
  \subfigure{
    \includegraphics[width=0.48\linewidth]{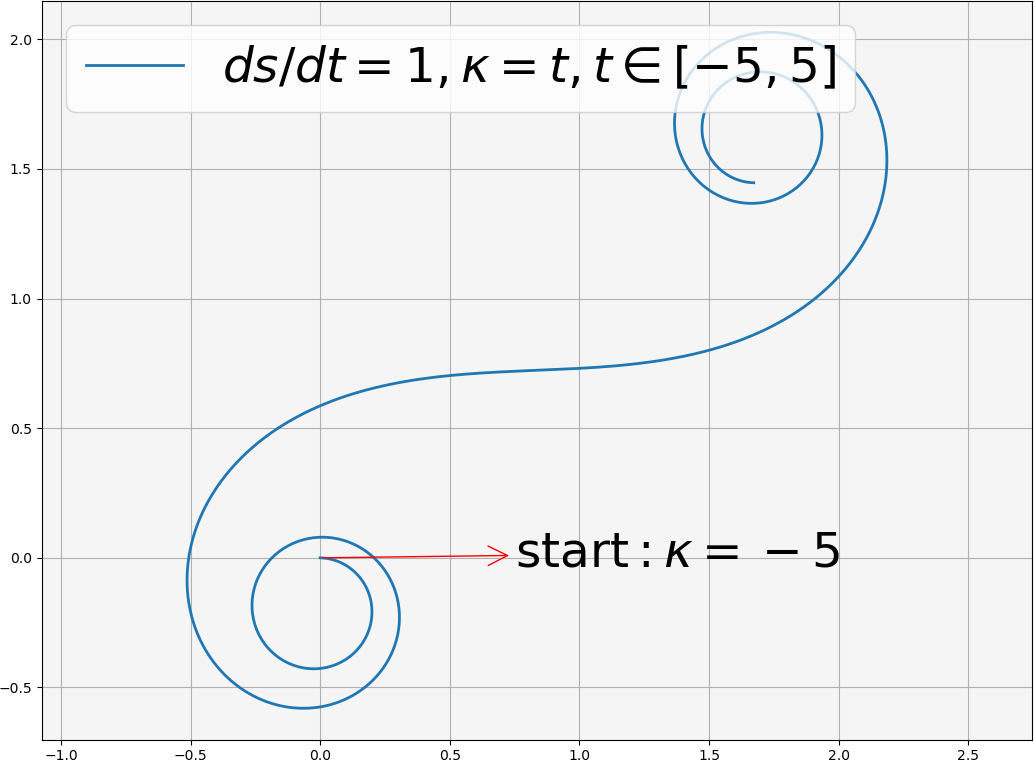}
  }
  \caption{The path generated with the proposed curvature planning model under different conditions.}
  \label{fig:fig2}
\end{figure}
The model of the curvature planning can be expressed as,
\begin{equation}
  \label{eq2}
  \begin{aligned}
     & \dot{\kappa}=\rho                                      \\
     & \dot{v}= a                                               \\
     & \dot{x}= \frac{ds}{dt}\cdot \cos \left( \theta \right) \\
     & \dot{y}= \frac{ds}{dt}\cdot \sin \left( \theta \right) \\
  \end{aligned}
\end{equation}
where $\rho$ represents the rate change of curvature, $a$ represents the pseudo acceleration along the curve.
\par
Consider $ds/dt$ to be the constant $1$, the waypoints can be strictly expressed as trigonometric functions with respect to the curvature, and the Eq.~\eqref{eq2} can be expressed as,
\begin{equation}
  \label{eq3}
  \left\{
  \begin{aligned}
     & \dot{\kappa}=\rho                   \\
     & \dot{\theta}=\kappa                 \\
     & \dot{x}= \cos \left( \theta \right) \\
     & \dot{y}= \sin \left( \theta \right) \\
  \end{aligned}
  \right.
\end{equation}
%

In the process of path planning, collision avoidance can be achieved by adjusting the curvature $\kappa$ in Eq.~\eqref{eq3}. In this way, the collision-free path generated by the FDSPC algorithm can be guaranteed to have $G^2$-continuity. The path generated with the proposed curvature planning method under different conditions are demonstrated in Fig.~\ref{fig:fig2}. 

\par
\subsection{Curvature planning in $2.5$-D terrain space}
In complex $3$-D environment, the curve satisfies the continuity of curvature and torsion \cite{guggenheimer2012differential}, must be twice differentiable in all directions. The Frenet-Serret equations are used to describe the $3$-D curve in the Frenet-Serret coordinate system. However, this method is relatively complex for pathfinding and is unnecessary for terrain-based mobile robots. In this work, we simplify the motion planning of $3$-D space to $2.5$-D path planning, which combines a continuous transformation scaling function $\tau$ in the $z$-direction and a $2$-D plane planning. The continuous variation of the curvature function $\kappa$ and the scaling function $\tau$ yields a $2.5$-D path that satisfies $G^2$ continuity.
\par
The curvature planning model in $2.5$-D space can be expressed as,
\begin{equation}
  \label{eq4}
  \left\{
  \begin{aligned}
     & \dot{\kappa}=\rho                   \\
     & \dot{\tau}_z=\rho _z                \\
     & \dot{\theta}=\kappa                 \\
     & \dot{x}= \cos \left( \theta \right) \\
     & \dot{y}= \sin \left( \theta \right) \\
     & \dot{z}=\tau _z                     \\
  \end{aligned}
  \right.
\end{equation}
where $\rho$ represents the rate of change of  the curvature, $\rho_z$ and  $\tau_z$ denote the rate of change and scaling function along the $z$-direction, respectively.\par
\begin{figure}[t]
  \centering
  \subfigure{
    \includegraphics[width=0.98\linewidth]{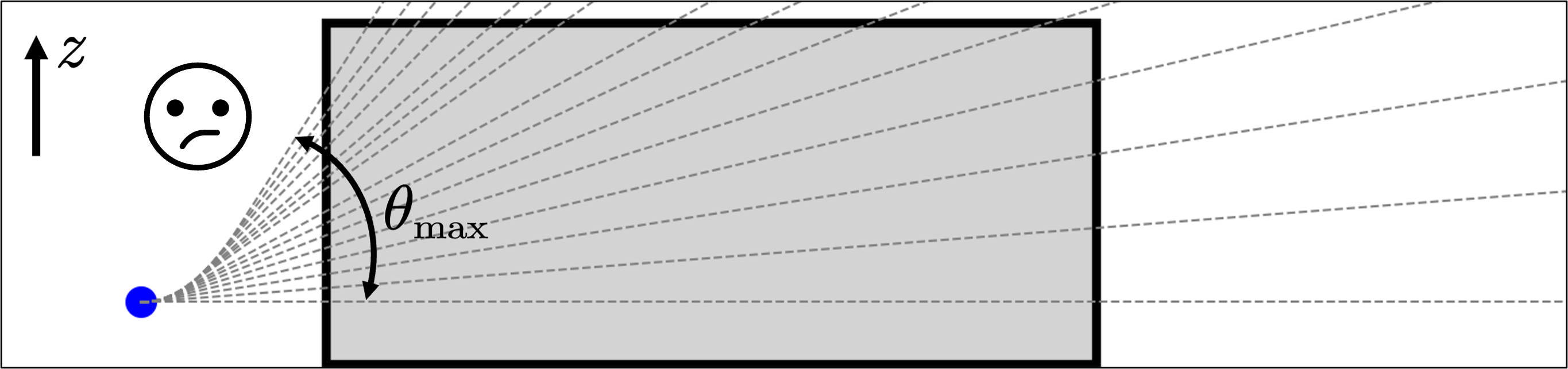}
    \label{figz:a}
  }
  \hspace{-5.5mm} 
  \subfigure{
    \includegraphics[width=0.98\linewidth]{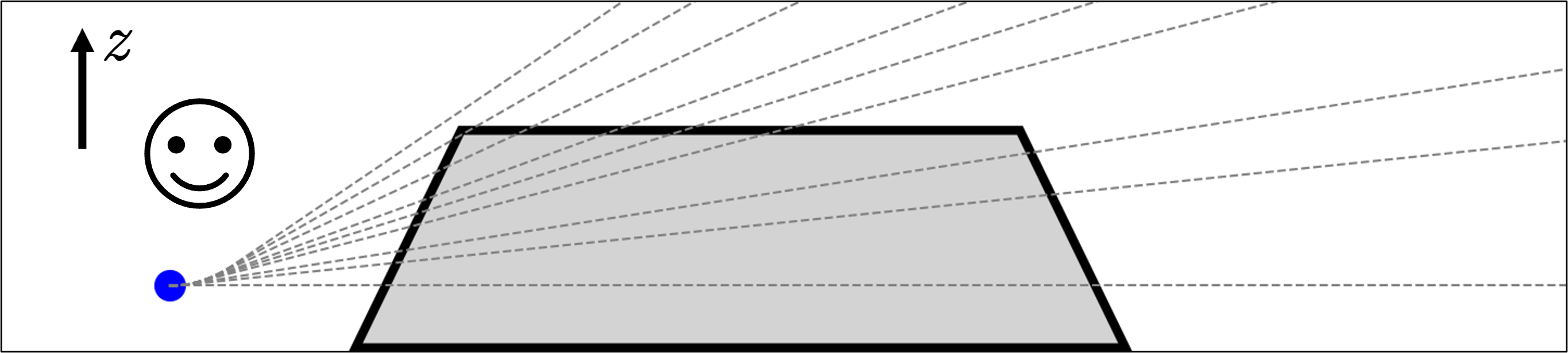}
    \label{figz:b}
  }
  \caption{The demonstration of $z$-axis expansion of the FDSPC when hitting an obstacle.}
  \label{figz}
\end{figure}
Once an obstacle is detected on the plane, FDSPC retreats a distance $back_{obs}$ and introduces a tilt along the $z$-axis. By comparing the $(x, y)$ coordinates of the new collision point with the original, it determines if the obstacle is crossable. If the $z$-tilt is below $\theta_{max}$, the algorithm continues $z$-axis expansion until a collision-free node is found; otherwise, it deems the obstacle uncrossable and backtracks. Two $z$-axis exploration cases are illustrated in Fig.~\ref{figz}.
\par

In practice, the mechanical structural and kinematics of a robot should be taken into consideration. For the path planning of the proposed wheel-legged robot, it might be favourable to keep the robot's torso height and posture stable, i.e., $\rho_z$ and $\tau_z$ in Eq.~\eqref{eq4} equal to $0$. 

\section{FDSPC ALGORITHM}
\subsection{Data structure}
\begin{figure}
  \centering
  \includegraphics[width=0.98\linewidth]{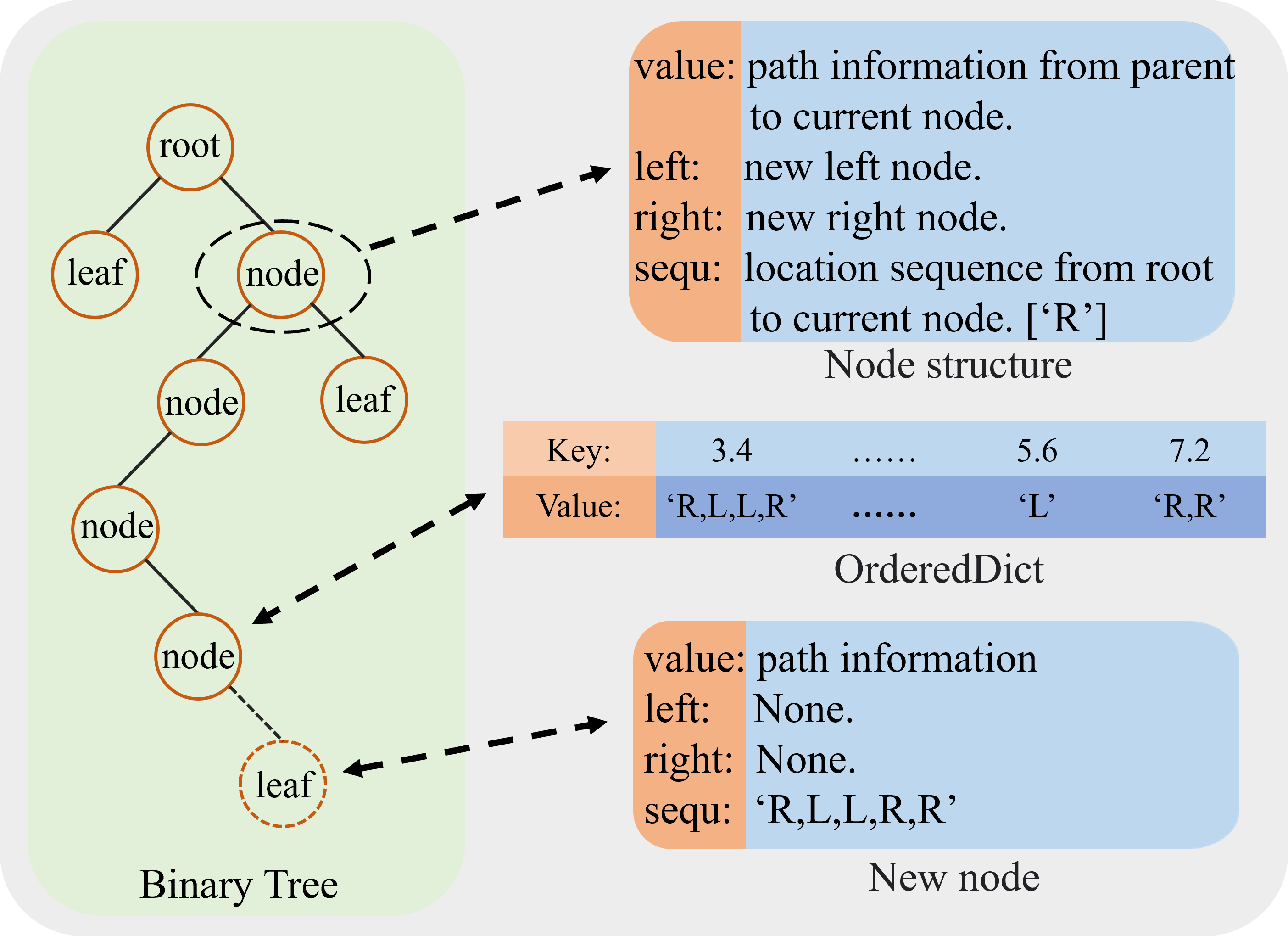}
  \caption{The data structure of the FDSPC algorithm.}
  \label{fig3}
\end{figure}

The path search process can be regarded as the construction of a binary tree from the root node (initial position) to the leaf nodes (target position). Each binary tree node stores path from the previous node to the current node. The node's search weight (the Euclidean distance from the current node position to the end point position) and the node's location (the node's position in the binary tree) is stored in an ordered dictionary. Whenever there arises a necessity to update and broaden potential path nodes, the location of the node with the lowest weight is popped out from the ordered dictionary, enabling a fast heuristic search of paths. The data structure of the FDSPC is illustrated in Fig.~\ref{fig3}.
\par
The path begins with the initialization of the root node in the binary tree, the destination is marked as the leaf node. Each node holds the $value$, representing the path from the preceding node to the current node. Additionally, the sequence stored in the $sequ$ attribute indicates the locating sequence in the binary tree from the root node to the current node. A corresponding node is created when a new branch point needs to be added. During expansion, the Euclidean distance from the current position to the end position is used as the $key$ (also regarded as the weight for node selection), and the locating sequence from the root node to the current node of the binary tree is viewed as the $value$. These are paired and added to the ordered dictionary. After adding a new node, prune the parent nodes in the ordered dictionary where both the left and right child nodes exist, indicating the infeasibility of further expansions.
\subsection{Smooth path planning on $2$-D plane}
\begin{algorithm}[t]
  \caption{Direct planning $\boldsymbol{D}_p(\cdot)$}\label{direct planning}
  Initialize $\theta_{g} \leftarrow \arctan2(p_{start},p_{goal})$;
  \par
  \If {$\theta_{g}\ne \theta_{0}$} {
    \While{True}
    {
      $\kappa_{s_1} \leftarrow \int{\rho}$; \\
      \tcp*[h]{Curvature integration}
      \par
      $\theta \leftarrow \int{\kappa_{s_{1}}}$;\\ 
      update $p_{now}$;\\
      $\theta_{g} \leftarrow \arctan2(p_{now},p_{goal})$;
      \par
      \If{$\theta \approx \theta_{g}$}{break;}
    }
  }
  \par
  $\kappa_{s_{2}} \leftarrow \mathbf{0}$;
  \par
  $\kappa_d \leftarrow \kappa_{s_{1}}+\kappa_{s_{2}}$;
  \par
  return $\kappa_d$;
\end{algorithm}
\par
The direct planning part of the FDSPC is illustrated in Alg.~\ref{direct planning}. Firstly, the algorithm checks whether the current orientation can reach the target position directly (i.e., if there are no obstacles along a straight line from current position to the target). If not, it adjusts the current orientation $\theta$ iteratively by using the curvature integral until the direction $\theta_g$ of the line connecting the current position and the target aligns with the $\theta$. Then, the current position $P_{now}$ is updated with $\theta$ by Eq.~\eqref{eq3}. By assigning a curvature value of $0$ to the straight-line segment, the algorithm sequentially connects the current position to the target. Essentially, the FDSPC constructs the basic path structure by combining segments of indefinite radius arcs with straight lines. The exploration planning part of the FDSPC is illustrated in Alg.~\ref{Explore planning}.
\par
\begin{algorithm}
  \caption{Explore planning $\boldsymbol{E}_{\boldsymbol{p}}(\cdot) $}\label{Explore planning}
  Initialize $\theta_t, \theta_0 \gets \arctan2(p_{now},p_{goal})$\;
  Initialize $\kappa_s \gets \boldsymbol{0}$\;
  Initialize $o_{idx} \leftarrow \boldsymbol{O}_d(\mathcal{F}_{map}, \kappa_s)$\;
  \While{$o_{idx}$ is not empty and $\left| \theta _t - \theta_0 \right| < \pi$}{
    \While{$o_{idx}$ is not empty}
    {
      $\kappa_{old} = \kappa_s$, $o_{idx,old} = o_{idx}$\;
      $l_{int} = o_{idx} + l_{add}$\;
      $\theta _t=\theta _t\pm \theta _{a_{1}}$\;
      $\kappa_s \gets \int_{inv}{\left( \theta _t,\rho ,len_{int} \right)}$\;
      $o_{idx} \leftarrow \boldsymbol{O}_d(\mathcal{F}_{map}, \kappa)$\;
    }
    \If{$\theta _{a_{1}} == \theta_{a_{2}}$}{
      break\;
    }
    $\theta _t=\theta _t\mp \theta _{a_{1}}$\;
    $\kappa_s = \kappa_{old}$\;
    $\theta _{a_{1}} = \theta _{a_{2}}$\;
  }
  $\kappa_a \gets {P_{fimin}}(\kappa_{old}[o_{idx,\,old}], \kappa_s)$\;
  return $\kappa_a$\;
\end{algorithm}
\par
Initially, $\boldsymbol{E}_{\boldsymbol{p}}(\cdot)$ establishes a connection between the current position and the endpoint along the orientation $\theta$ using a straight line, and calculates the position index of the collision. If the index is non-empty and the extension angle is less than $\pi$, it increases the extension angle $\theta_{t}$ on both sides of the obstacle progressively until the index of the collision become empty. The length of integration exploration $l_{int}$ is adjusted based on the position of collision index $o_{idx}$ and the linear extension length $l_{add}$ to avoid overly long exploration distances that might mistakenly identify boundaries as obstacles. In line 9 of Alg. \ref{Explore planning}, a sequence $\kappa_s$ of curvatures is obtained through inverse integration under specified integration length $l_{int}$, integral value $\theta_t$, and integration rate $\rho$. Upon the program runs into the inner while loop (line 5 to line 11, Alg. 2) for the second time, the extension angle begins increasing from the previous collision angle by increments $\theta_{a_2}$ ($\theta_{a_2}<\theta_{a_1}$) to refine the exploration of collision edges until the position index of the collision point is empty. Finally, the algorithm utilizes the ${P_{fimin}}(\cdot)$ function to locate a point $p$ on the current collision-free curvature path $\kappa_s$ that minimizes the distance to the previous collision position on the curvature path $\kappa_{old}$. Subsequently, the algorithm returns the collision-free curvature path $\kappa_p$ preceding point $p$ on $\kappa_s$.
\par
In Alg. \ref{Explore planning}, if the extension angle to either side, i.e., $\left| \theta _t - \theta_0 \right|$, exceeds $\pi$, the exploration node is closed. This process involves pruning the extra nodes, and then popping a new node from the binary tree $\mathcal{B}_t$.
\par
\subsection{Smooth path planning in $2.5$-D space}
\begin{figure}[t]
  \centering
  \subfigure{
    \includegraphics[width=0.98\linewidth]{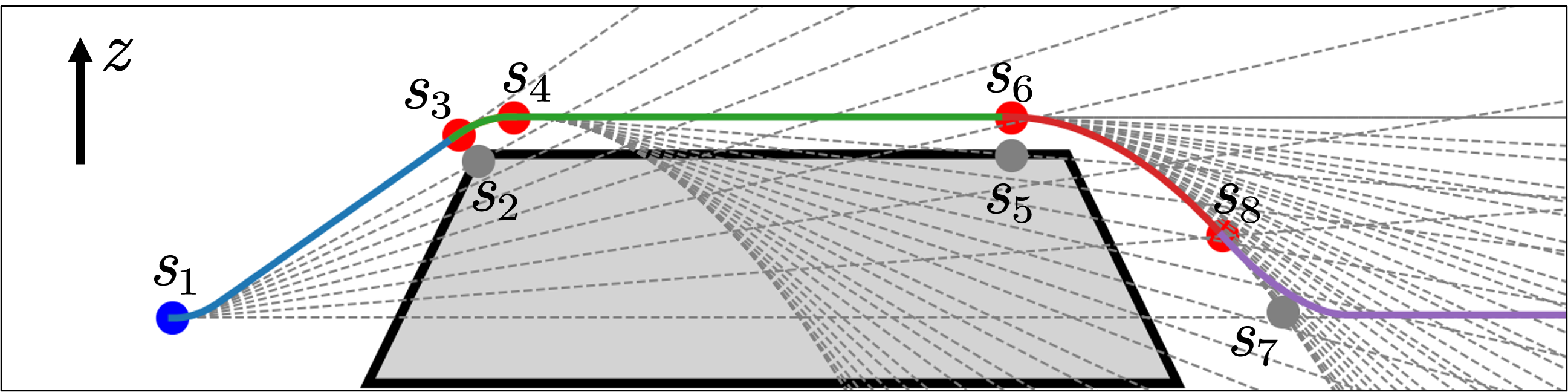}
    \label{zzz:a}
  }
  \hspace{-15mm} 
  \subfigure{
    \includegraphics[width=0.98\linewidth]{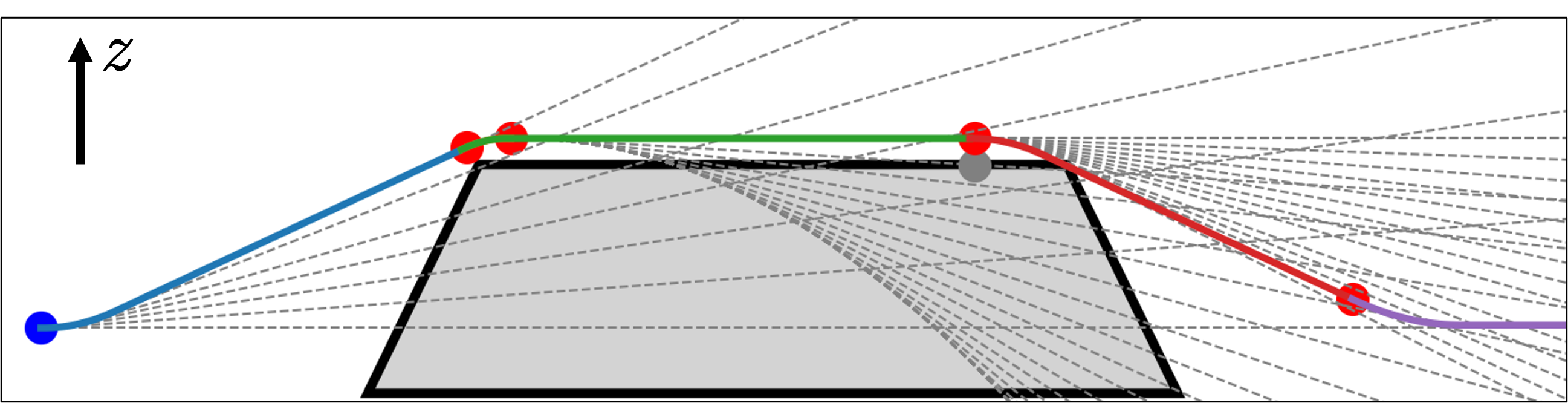}
    \label{zzz:b}
  }
  \caption{The demonstration in 2 different scenarios of $z$-axis expansion by the FDSPC when hitting an obstacle.}
  \label{zzz}
\end{figure}
As shown in Fig.~\ref{zzz}, if the obstacle on the plane is deemed crossable, the algorithm first identifies the point $s_3$ on the collision-free path that is closest to the previous point of collision $s_2$. It then generates a collision-free smooth path along the upward slope segment (the blue line). At point $s_3$, a negative $\rho_z$ is applied to transform the original path into a horizontal one (the green line). At point $s_4$, where the green path becomes horizontal, the algorithm searches for the point $s_6$ closest to the previous collision point $s_5$ on the first collision-free path from $\theta_{\textup{max}}$ to $0$. At point $s_6$, the algorithm searches for a ground collision point $s_7$ at a specified height, scanning from $0$ to $-\theta_{\textup{max}}$, and retreats $back_{obs}/2$ to $s_8$. At point $s_8$, a smooth path to the horizon is constructed (purple path), resulting in a smooth vertical path that crosses low obstacles.
\par
\begin{algorithm}[t]
  \caption{FDSPC}\label{overview}
  Initialize $\kappa_d \gets \boldsymbol{D}_p(p_{start},\, p_{goal}, \,\theta_\mathit{0})$\;
  Initialize $obs_{idx} \leftarrow \boldsymbol{O}_d(\mathcal{F}_{map}, \kappa_d)$\;
  update $p_{now}$, $\mathcal{B}_t(p_{now})$, $\mathcal{O}_d(p_{now})$\;

  \While{$obs_{idx}$ $\land$  $\mathcal{B}_t$ is not empty}
  {
    $\kappa_a \gets \boldsymbol{E}_p(\mathcal{F}_{map},\, p_{now},\, p_{goal})$\;
    update $p_{now}$, $\mathcal{B}_t(p_{now})$, $\mathcal{O}_d(p_{now})$\;
    $\kappa_d \gets \boldsymbol{D}_p(p_{start}, p_{goal}, \theta_\mathit{0})$\;
    $obs_{idx} \leftarrow \boldsymbol{O}_d(\mathcal{F}_{map}, \kappa_d)$\;
    update $p_{now}$, $\mathcal{B}_t(p_{now})$, $\mathcal{O}_d(p_{now})$\;
  }
  \eIf {$\mathcal{B}_t$ is empty}
  {
    \tcp*[h]{Find feasible path failed}\;
    return $NULL$\;}
  {
    $\kappa \leftarrow \mathcal{B}_t$\;
    return $\kappa$\;
  } 
\end{algorithm}
Alg. \ref{overview} outlines the framework of the FDSPC. Initially, the algorithm generates an initial curvature path $\kappa_d$ via the direct planning function $\boldsymbol{D}_p(\cdot)$. Subsequently, the collision detection function $\boldsymbol{O}_d(\cdot)$ to assess the feasibility of the path. If a collision is detected, the algorithm retraces its steps along the collision point sequence $back_{obs}$ to determine a branch point $b$. This branch point along with the collision-free branching path, is then incorporated into the binary tree $\mathcal{B}_t$ and the ordered dictionary $\mathcal{O}_d(\cdot)$. Using the exploration function $\boldsymbol{E}_p(\cdot)$, a short collision-free curvature path $\kappa_a$ is obtained. $\mathcal{B}_t$ and $\mathcal{O}_d(\cdot)$ are updated accordingly. The algorithm then iterates through the direct planning function $\boldsymbol{D}_p(\cdot)$ again to derive a new curvature path $\kappa_d$, repeating this cycle until a collision-free curvature path $\kappa_d$ is found. Finally, the algorithm traces back through the branches of the binary tree $\mathcal{B}_t$ from the endpoint leaf to obtain the final curvature path $\kappa$.
\par
The algorithm selects the position of the node with the lowest weight in the popped binary tree based on the $key$ of the ordered dictionary for subsequent expansion. For the planning near obstacles, the newly added node often coincides with the position of the last expansion node after updating the collision point, as depicted in the subfigure of Fig.~\ref{fig2:subfig:b}. This scenario poses a risk of generating pseudo-feasible paths, where all newly added paths are collision-free within the specified expansion length but do not reach the endpoint, potentially leading to premature termination of the algorithm. To mitigate this, for pseudo-feasible paths encountered in Alg. \ref{Explore planning}, we select a small segment and continue planning from the new node position. Essentially, FDSPC must begin and end with Direct planning $\boldsymbol{D}_p(\cdot)$, and with Explore planning $\boldsymbol{E}_p(\cdot)$ interspersed in between.
\par
\subsection{Velocity planning}
Since the path obtained by FDSPC is of $G^2$ continuity, the corresponding velocity and acceleration functions can be generated based on the curvature $\kappa$ or $\rho_z$, which ultimately yields a smooth trajectory. The velocity planning function except for the beginning and ending part is as follows,
\begin{equation}
  \begin{aligned}
    v_i = 
    \begin{cases}
      v_{i-1} + a \cdot dt, & \text{if } (\kappa_i = 0 \lor \rho_z = 0) \land \\
                            & \quad v_{i-1} < v_{\max} \\
      v_{i-1} - a \cdot dt, & \text{if } (\kappa_i \ne 0 \lor \rho_z \ne 0) \land \\
                            & \quad v_{\min} < v_{i-1} < v_{\max} \\
      v_{i-1},              & \text{otherwise}
    \end{cases}
  \end{aligned}
  \label{eq5}
\end{equation}
where $a$ denotes the acceleration, $\kappa_i$ is the curvature at the $i$-th point, $\rho_z$ is the curvature along the $z$-axis, $v_{\max}$ and $v_{\min}$ are the maximum and minimum velocities, respectively.
\par
Compared with other velocity planning methods, such as those based on path coordinate positions \cite{li2023apollo,6832481}, curvature-based velocity planning can anticipate changes in the path in advance, thereby adjusting the velocity in a timely manner to enhance the comfort of the vehicle and the stability of mobile robots. In contrast to optimization-based velocity planning methods \cite{cao2017optimal}, curvature-based velocity planning is simpler, faster, and does not require complex numerical optimization.
\par
\section{SIMULATION AND ANALYSIS}
In this section, the performance of FDSPC and some state-of-the-art methods \cite{pythonmotionplanning} are tested and compared five typical scenarios. Additionally, the proposed FDSPC method is also successfully applied on our self-designed wheel-legged robot to across a dike shaped obstacle.
\subsection{Simulation results}
\begin{figure}[h]
    \centering
    \includegraphics[width=0.98\linewidth]{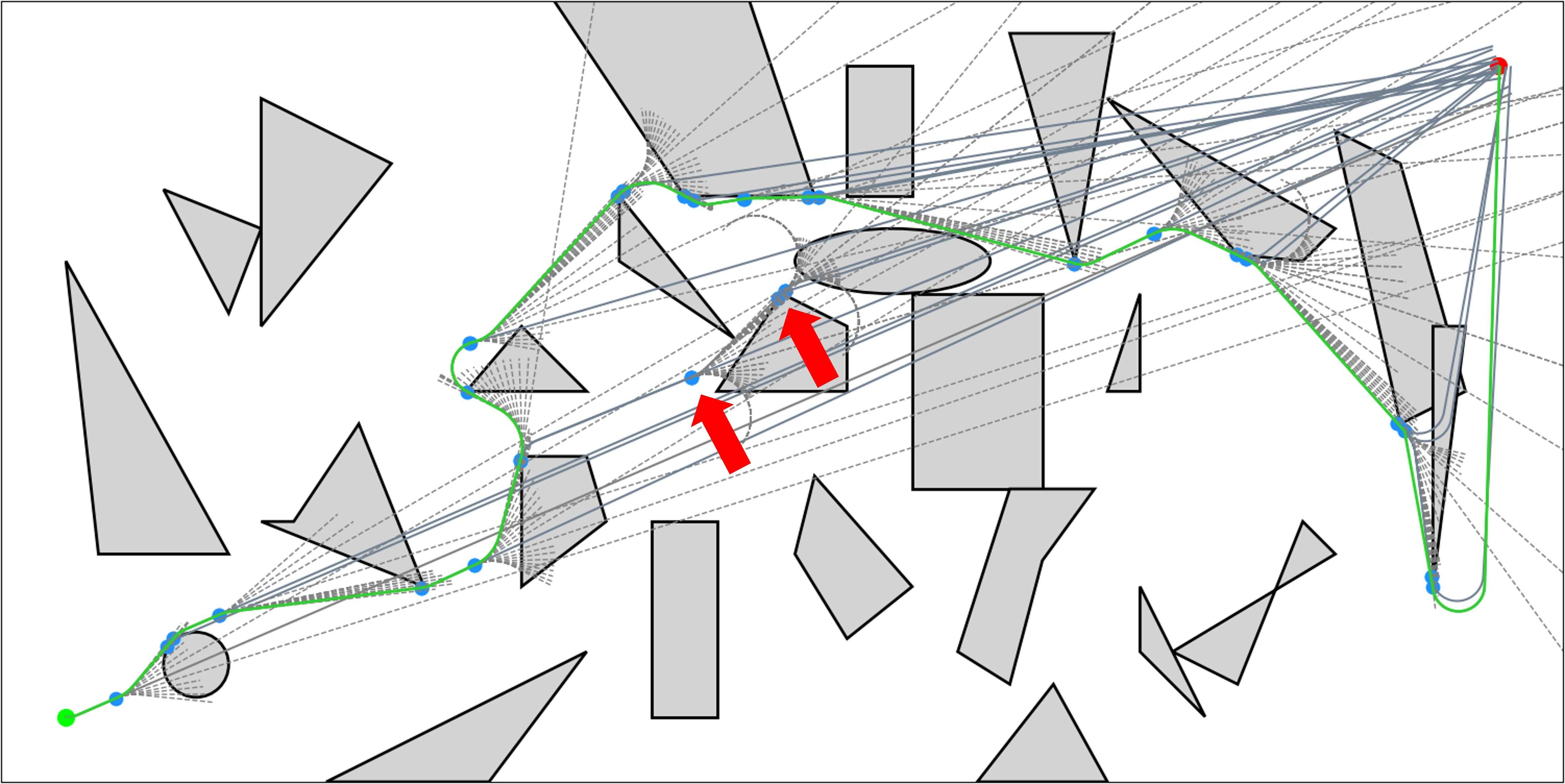}
    \caption{Simulation process and result of FDSPC algorithm in a randomly scattered polygonal map.}
    \label{complex2}
\end{figure}

\begin{figure}[h]
  \centering
  \subfigure[Long Obstacle]{
    \includegraphics[width=0.45\linewidth]{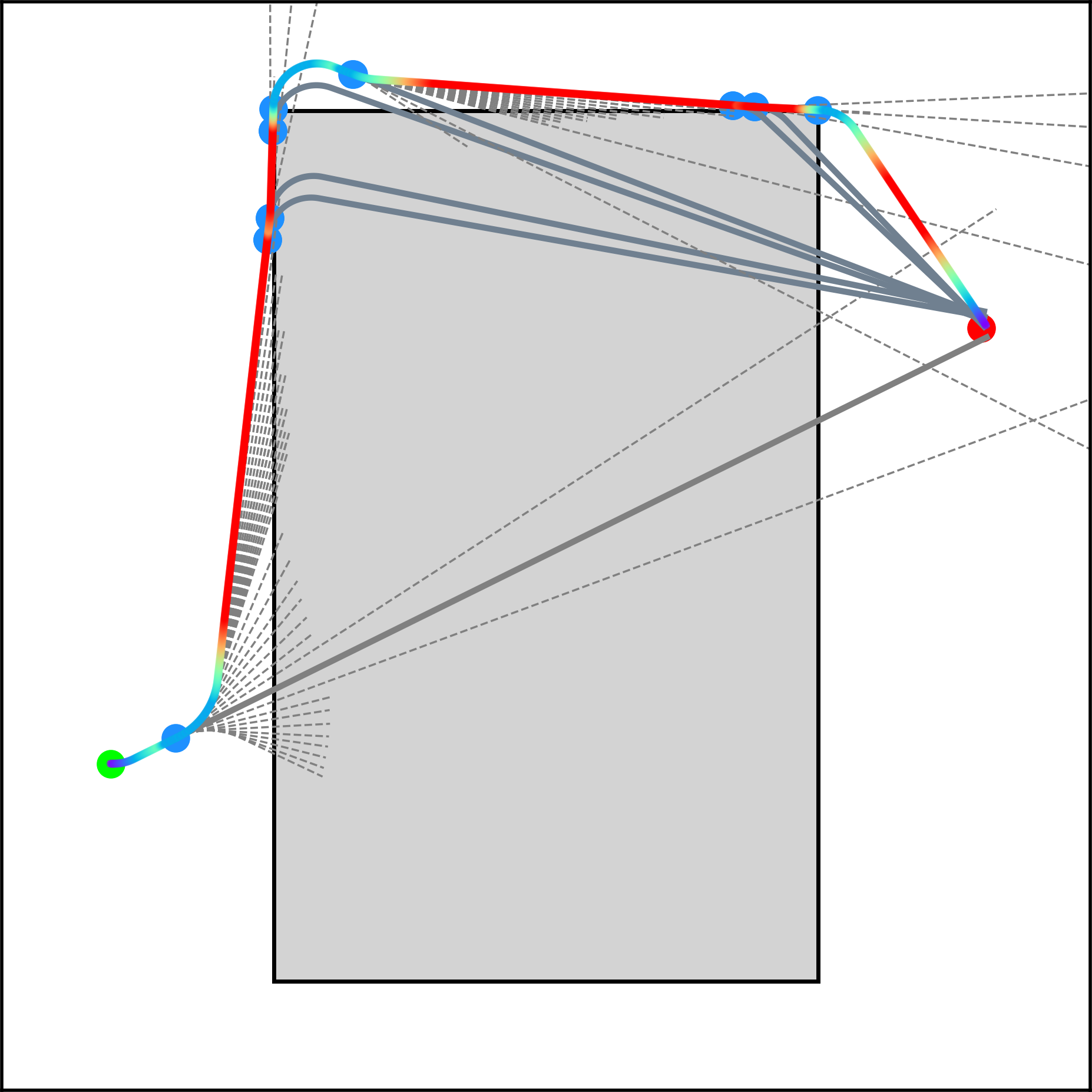}
    \label{fig2:subfig:a}
  }
  \subfigure[Long Corridor]{
    \includegraphics[width=0.45\linewidth]{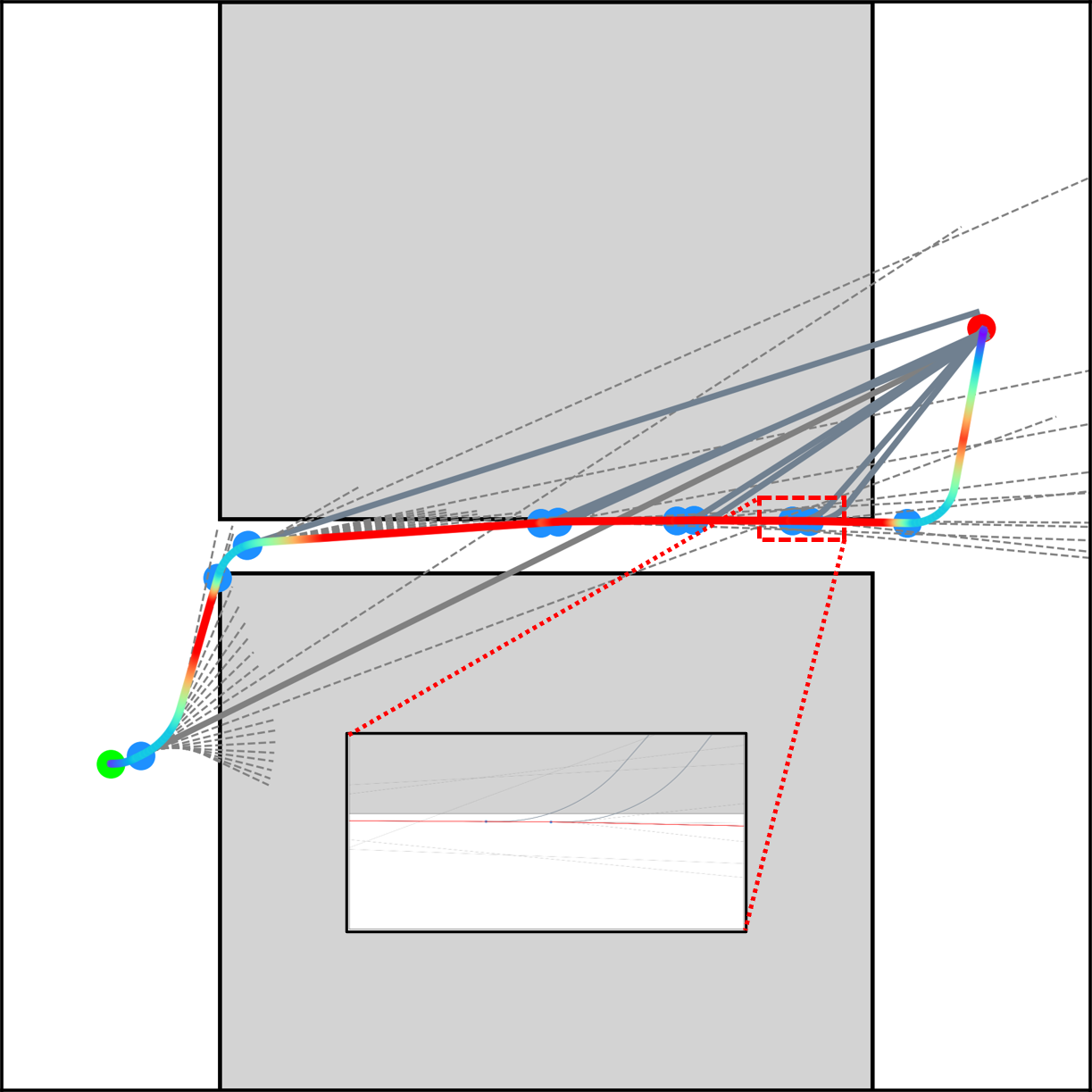}
    \label{fig2:subfig:b}
  }
  \subfigure[Semi-Enclosed]{
    \includegraphics[width=0.45\linewidth]{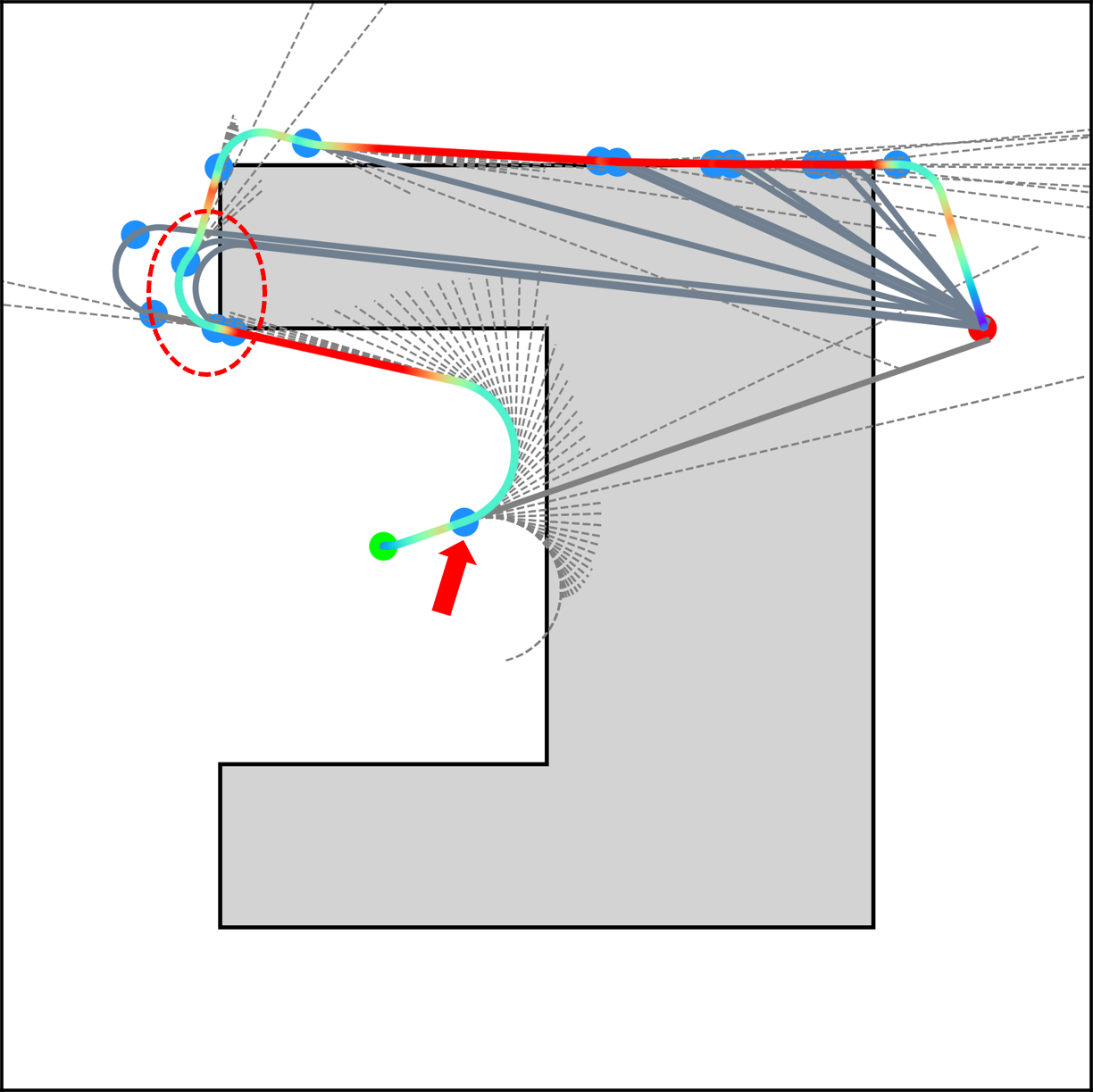}
    \label{fig2:subfig:c}
  }
  \subfigure[Random Complex]{
    \includegraphics[width=0.45\linewidth]{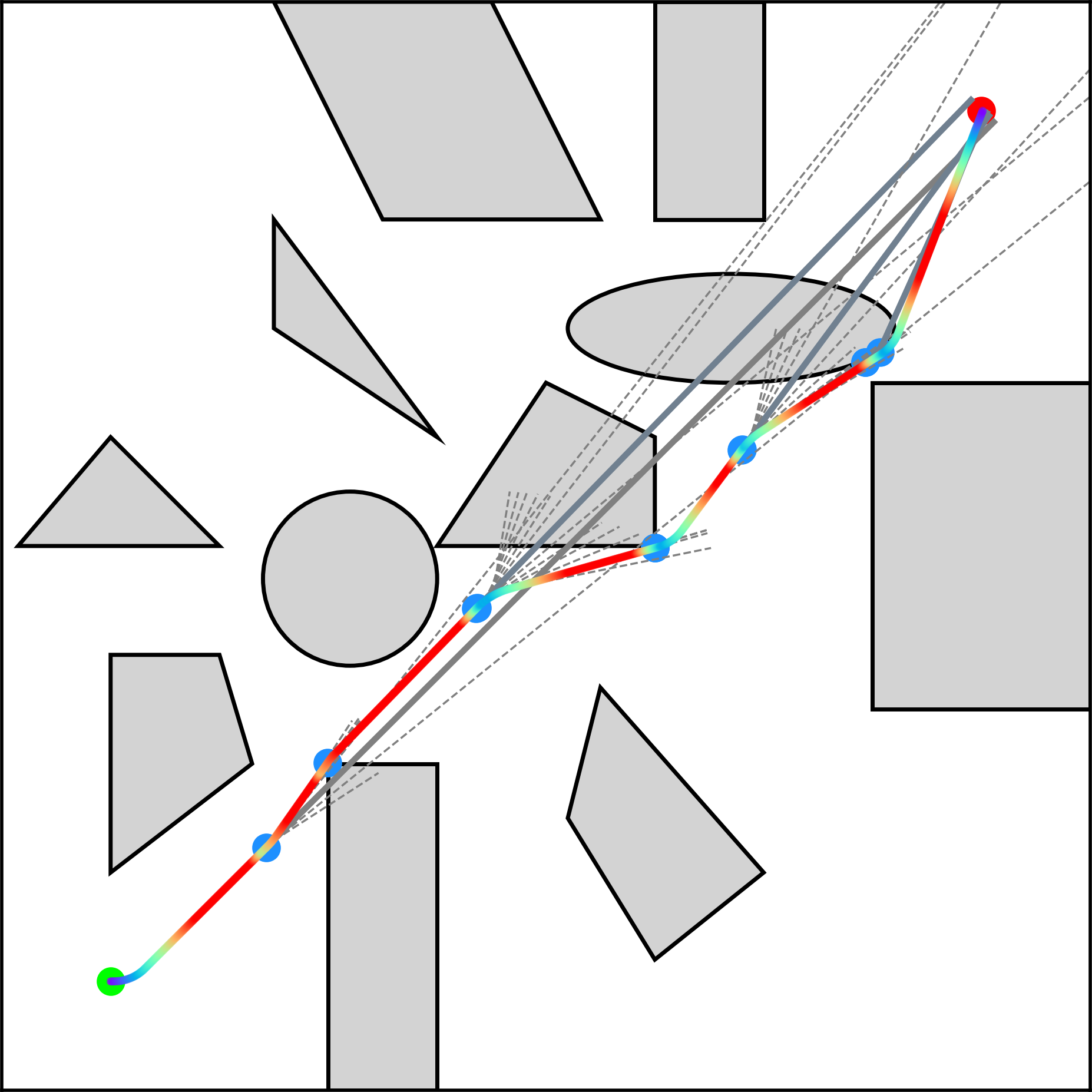}
    \label{fig2:subfig:d}
  }
  \subfigure[Simple Maze]{
    \includegraphics[width=0.95\linewidth]{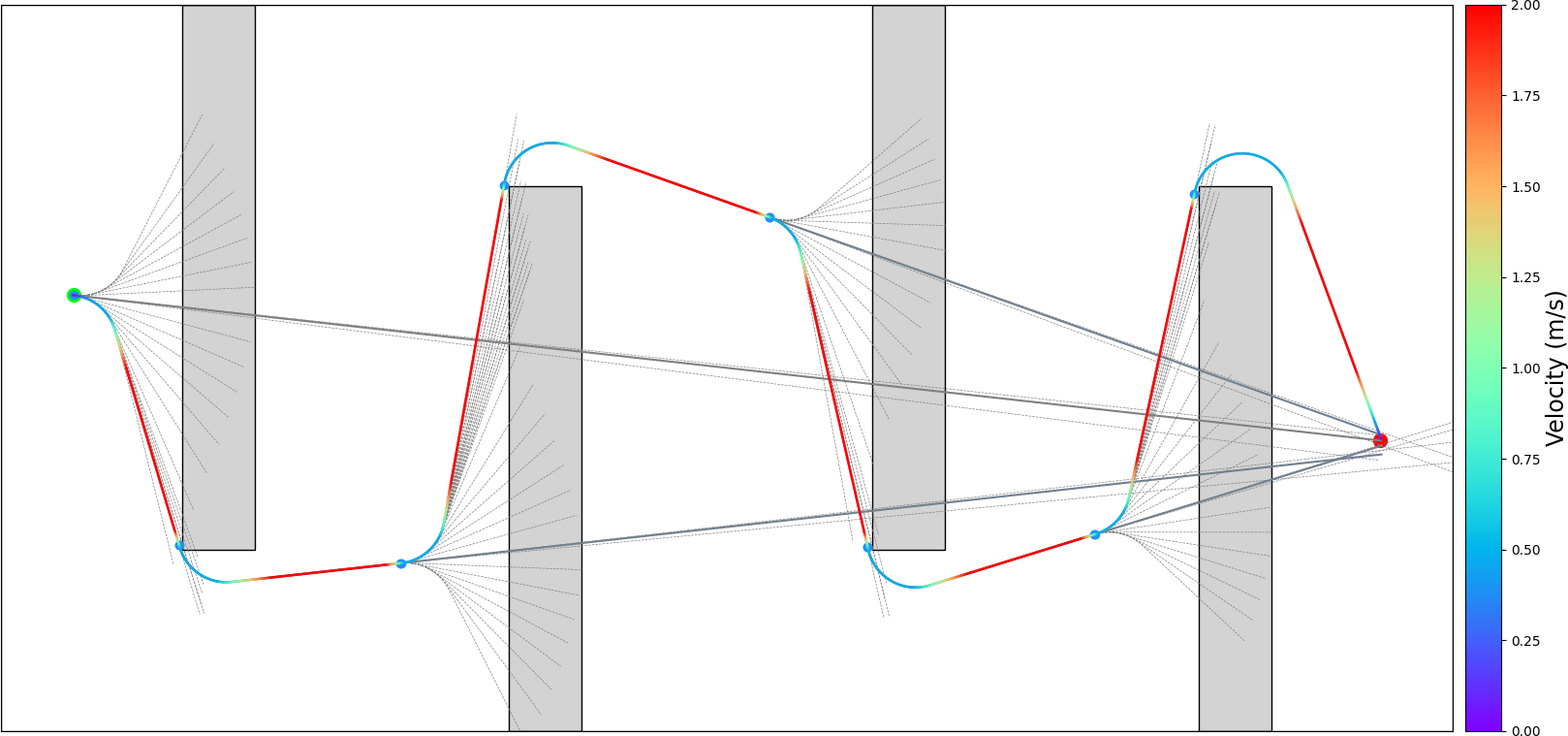}
    \label{fig2:subfig:e}
  }
  \caption{Simulation and results with adding velocity of FDSPC in five representative scenarios, where $\Delta t = 0.01$, $\rho = 0.4$, $\theta_{a1}=0.1$, $l_{add}=0.5$.}
  \label{fig2}
\end{figure}

In the validation, we constructed a randomly scattered polygonal map, as shown in Fig.~\ref{complex2}. During the path generation, a total of 27 $\mathcal{B}_t$ nodes (blue points) are generated, including 3 discarded nodes (marked by red arrows). Ultimately, a path satisfying $G^2$ continuity is generated. Additionally, we designed five typical scenarios, including bypassing long obstacles, navigating through long corridors and semi-enclosed areas, traversing random complex and simple maze environments. The specific exploration process and the final smooth path with velocity are shown in Fig.~\ref{fig2}.

\par
In the bypassing-long-obstacle scenario, FDSPC successfully identifies a short and smooth path by navigating close to the obstacle edge. As shown in Fig.~\ref{fig2:subfig:a}, \ref{fig2:subfig:b}, \ref{fig2:subfig:d}, and \ref{fig2:subfig:e}, aided by the heuristic function, FDSPC efficiently finds $G^2$-continuous paths in long corridor, random complex, and simple maze scenarios. In the semi-enclosed scenario (Fig.~\ref{fig2:subfig:c}), FDSPC explores both sides of the obstacle and determines that the shortest feasible path passes on the left, adding collision-free branch nodes to the binary tree (red circle). When a node is popped from the ordered dictionary, the branch in the red circle is farther from the goal and thus weighted higher. The algorithm prioritizes expansion from the unvisited (right) direction of the earlier node (red arrow). If the expansion angle exceeds $\pi$ and no valid node is found, the branch is pruned and planning continues from the next node. Across scenarios, pseudo-feasible paths are sometimes encountered. By gradually extending these paths (e.g., two adjacent blue points in Fig.~\ref{fig2:subfig:b}, \ref{fig2:subfig:c}), FDSPC ultimately identifies smooth, feasible solutions.
\par

\subsection{Performance analysis}
\begin{figure*}[htp]
  \centering
    \includegraphics[width=0.197\linewidth]{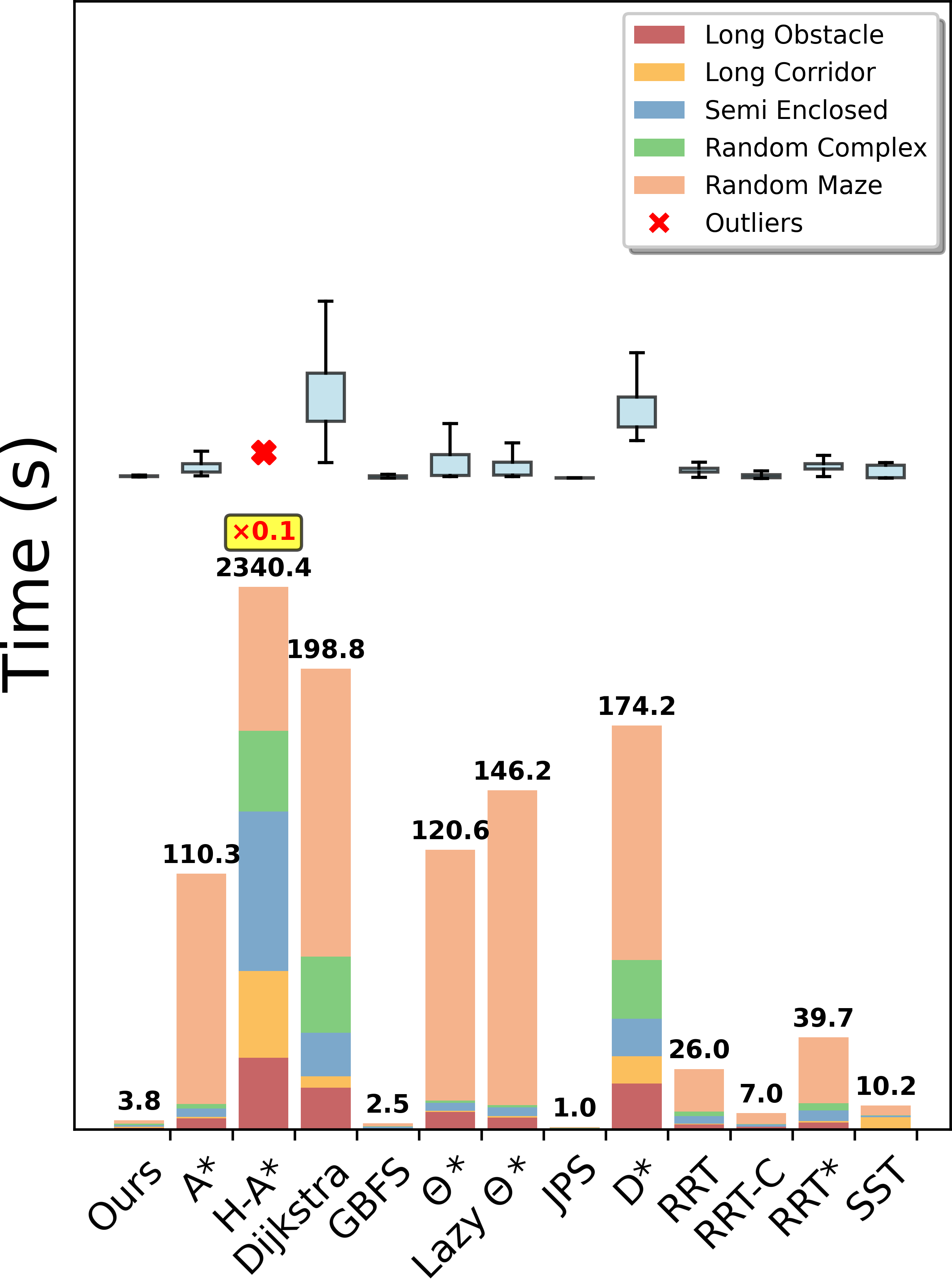}
  \hspace{-2mm} 
    \includegraphics[width=0.197\linewidth]{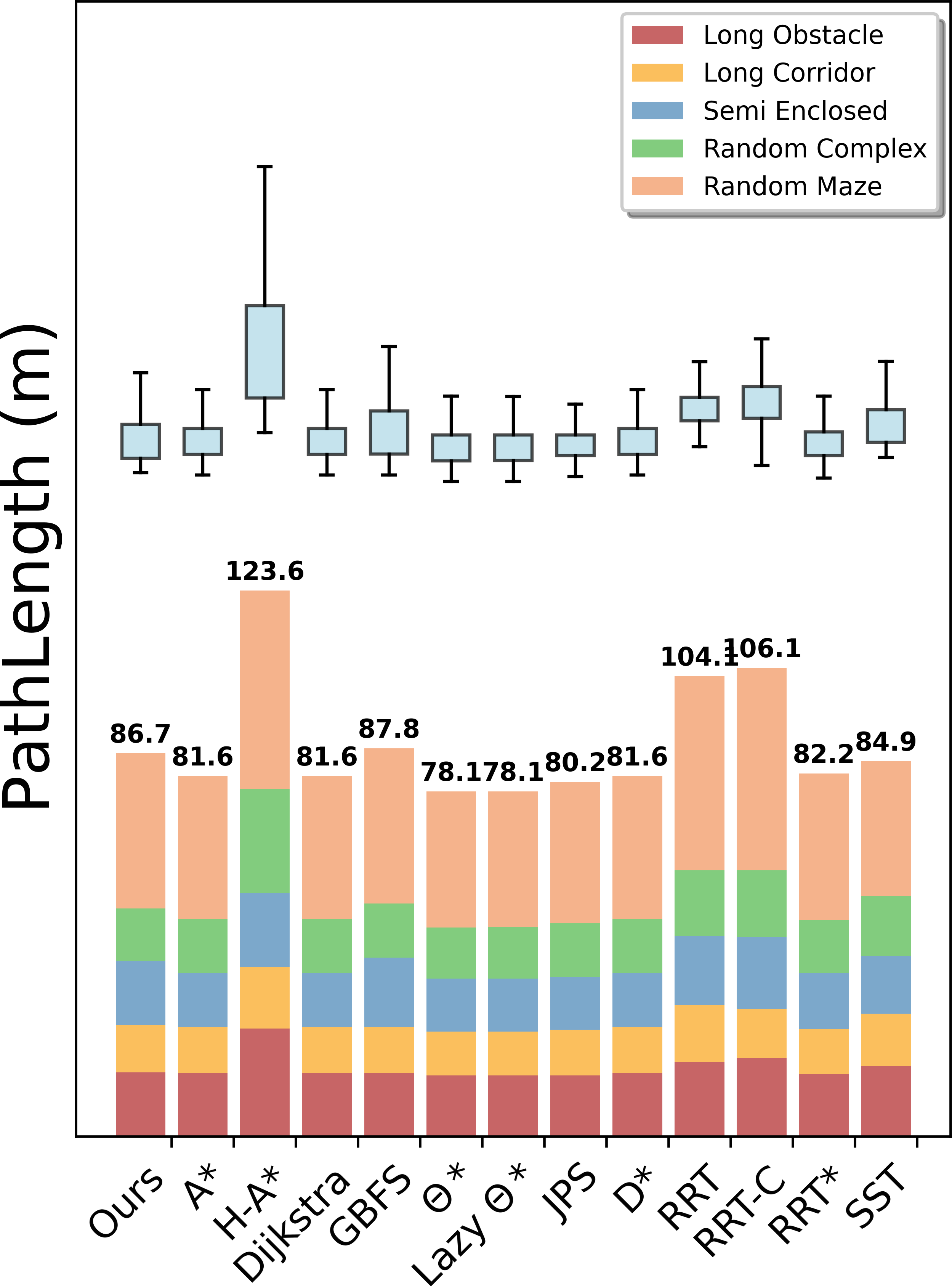}
  \hspace{-2mm} 
    \includegraphics[width=0.197\linewidth]{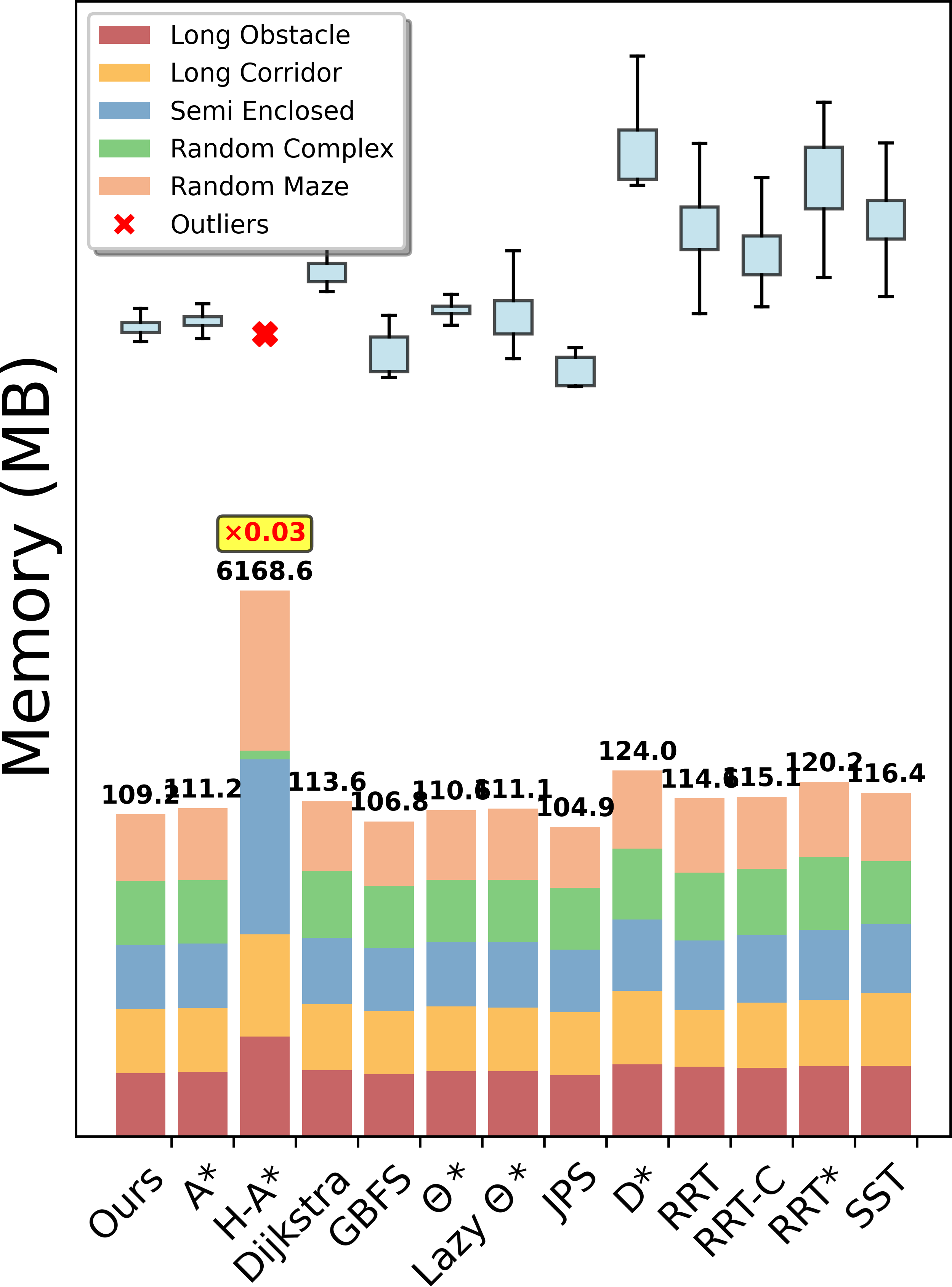}
  \hspace{-2mm} 
    \includegraphics[width=0.197\linewidth]{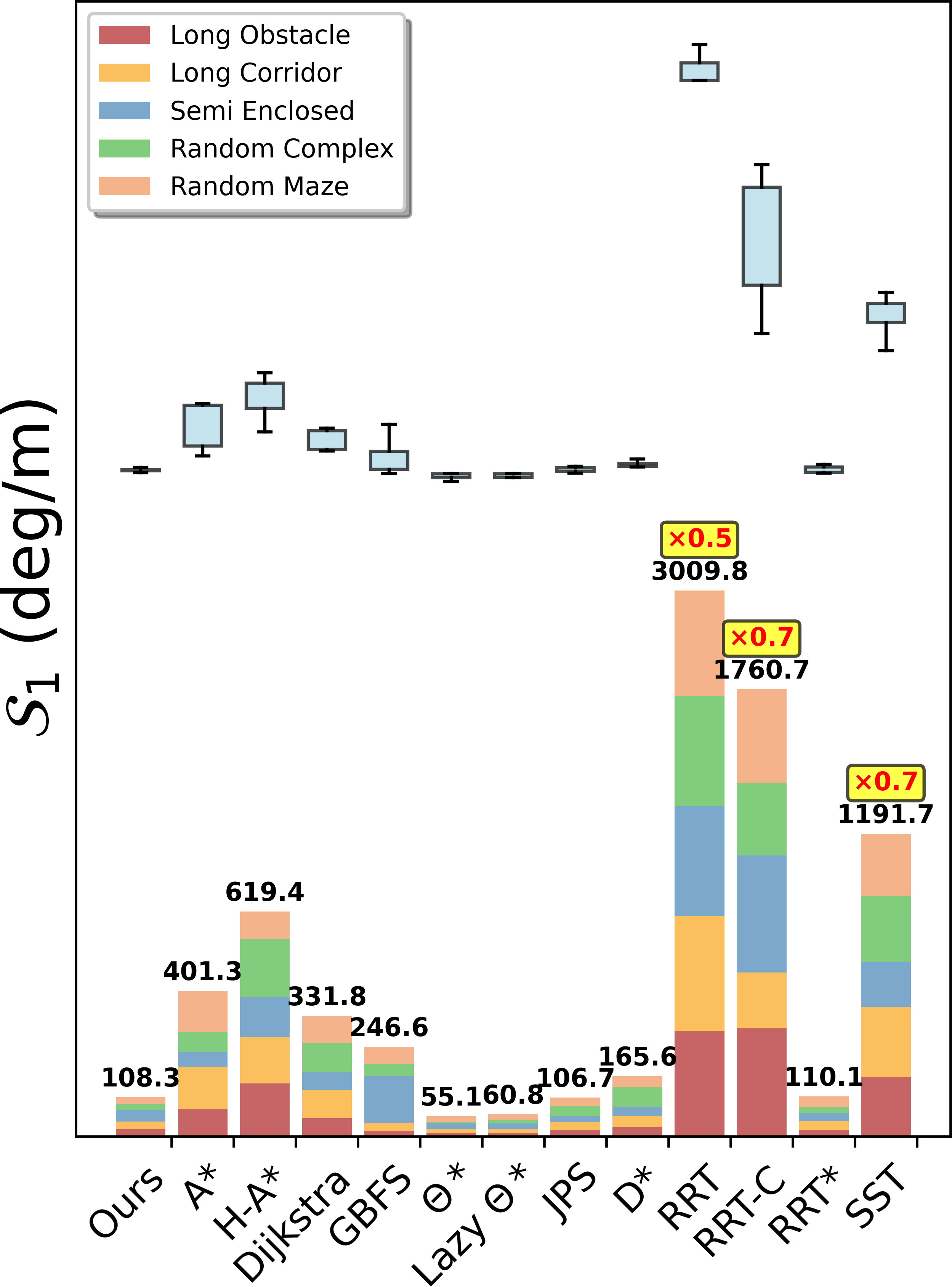}
  \hspace{-2mm} 
    \includegraphics[width=0.197\linewidth]{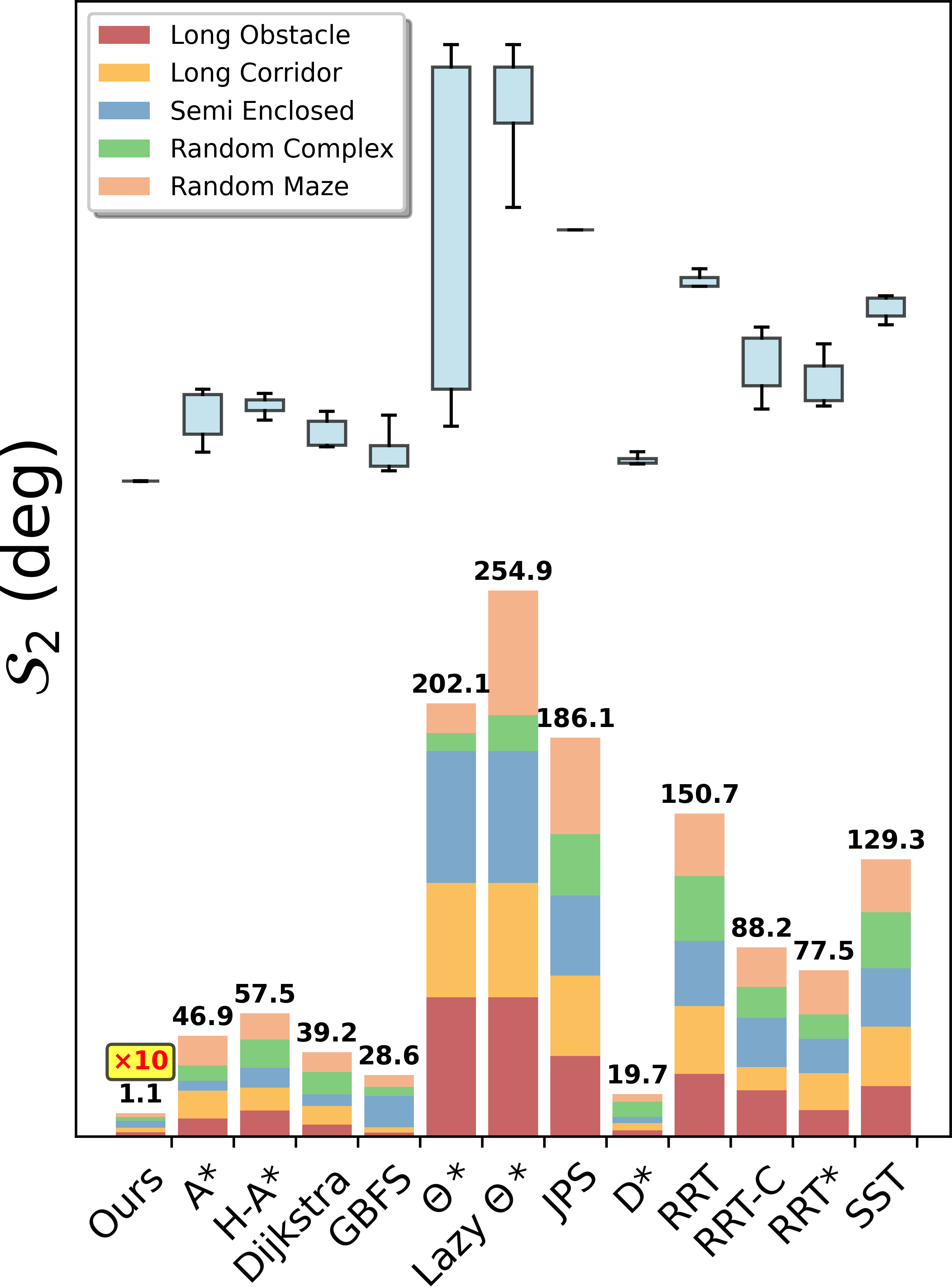}
  \caption{Comparative analysis of path planning algorithms across five performance metrics and five scenarios. The box plot in the upper part shows the performance of various algorithms in five scenarios, and the lower part is a bar chart composed of specific numerical values. Some of the bar charts have been scaled (yellow magnification factor) to better present the comparison among the algorithms.}
  \label{fig_smooth2}
\end{figure*}

The performance of the FDSPC algorithm is compared with other algorithms in terms of path-solving time, memory usage, path length and smoothness (measured by $\mathcal{S}_1$ and $\mathcal{S}_2$). The results are presented in Table\ref{performance} and Fig.~\ref{fig_smooth2}, with path smoothness calculated using Eq.~\eqref{eq:G1}. $\mathcal{S}_1$ and $\mathcal{S}_2$ are defined as the average change in heading per unit length along the path and mean angle of turning, as follows,
\begin{equation}
  \label{eq:G1}
  \begin{aligned}
    \mathcal{S} _1 & = \frac{\sum_{i=1}^{n-1}{\theta _i}}{l_p}  ,\,\,\,\,\,
    \mathcal{S} _2  = \frac{\sum_{i=1}^{n-1}{\theta _i}}{l_c}  \\
      \theta_i &= \begin{cases}
      \arccos \left( \frac{\mathbf{v}_{i-1} \cdot \mathbf{v}_i}{|\mathbf{v}_{i-1}| |\mathbf{v}_i|} \right), & \text{if } |\mathbf{v}_{i-1}| \neq 0 \land  |\mathbf{v}_i| \neq 0, \\
      0,  \text{otherwise,}
      \end{cases}
  \end{aligned}
\end{equation}
where $l_p$ is the path length, $l_c$ is the number of $\theta_i$ that is not 0, $n$ is the number of path points, and $\mathbf{v}_{i-1} = p_i - p_{i-1}, \quad \mathbf{v}_i = p_{i+1} - p_i$, $p_i$ represents the point coordinate of the path.
\par
\begin{table*}[ht]
  \centering
  \caption{The performance comparison of the FDSPC algorithm with state-of-the-art methods in five typical scenarios.}
  \setlength{\tabcolsep}{4.5pt} 
  \begin{tabular}{@{}clcccccccccccccc@{}}
    \toprule
    \multicolumn{2}{c}{ }    & Ours      & A*  & Hybrid-A*  & Dijkstra & GBFS   & $\theta$*    & lazy $\theta$* & JPS    & D*    & RRT    & RRT-C  & RRT* & SST  & Mean     \\
    \midrule
    \multirow{5}{*}{\makecell{Long \\ Obstacle}}
           & Time               & 0.55  & 4.61  & 308.27  & 17.81  & $\pmb{0.15}$ & 7.43           & 5.07   & 0.18  & 19.67  & 2.04    & 1.11   & 2.93 & 0.27 & 5.61   \\
           & Memory             & 21.47 & 21.74 & 1128.29  & 22.74  & 20.96        & 22.06          & 22.08  & $\pmb{20.73}$ & 24.33  & 23.59  & 23.23  & 23.68 & 23.82 & 22.70  \\
                                    & Length             & 14.48 & 14.30  & 24.39  & 14.30   & 14.30        & $\pmb{13.76}$  & $\pmb{13.76}$  & $\pmb{13.76}$ & 14.30   & 16.85  & 17.75  & 14.02 & 15.82 & 16.13  \\
                                    & $\mathcal{S}_1 $   & 19.38 & 75.52 & 145.79  & 5035   & 15.73        & $\pmb{9.45}$           &  $\pmb{9.45}$  & 16.35 & 25.17  & 581.85 & 428.04 & 17.22  & 233.16 & 143.37 \\
                                    & $\mathcal{S}_2 $   & $\pmb{0.19}$  & 8.31  & 12.06  & 5.54   & 1.73         & 65.00             & 65.00     & 37.50  & 2.77   & 29.13  & 21.43  & 12.16   & 23.39 & 19.93  \\
    \midrule
    \multirow{5}{*}{\makecell{Long \\ Corridor}}  & Time                  & 0.46  & 0.79  & 374.03 & 4.95   & 0.17         & 0.54           & 0.56   & 0.15  & 11.89  & 0.39   & $\pmb{0.03} $  & 0.61  & 4.92  & 2.31   \\
                                    & Memory                              & 21.62 & 21.77 & 1155.21  & 22.30   & 21.49        & 21.93          & 21.59  & 21.37 & 25.00     & $\pmb{19.14 }$  & 22.05  & 22.54  & 24.85 & 22.41  \\
                                    & Length                              & 10.66 & 10.48  & 13.96 & 10.48  & 10.48        & $\pmb{9.96}$   &  $\pmb{9.96}$ & 10.36 & 10.48  & 12.79  & 11.14  & 10.23  & 11.93 & 10.82  \\
                                    & $\mathcal{S}_1 $                    & 21.32 & 115.97 & 127.87 & 77.31  & 21.48        & $\pmb{10.71}$  & $\pmb{10.71}$  & 21.72 & 30.07  & 633.38 & 217.17 & 24.78  & 276.55 & 85.09 \\
                                    & $\mathcal{S}_2 $                    & $\pmb{0.21}$  & 13.06  & 10.69 & 8.71   & 2.42         & 53.33          & 53.33  & 37.50  & 3.39   & 31.73  & 10.88  & 17.31  & 27.73 & 19.70  \\
    \midrule
    \multirow{5}{*}{\makecell{Semi \\ Enclosed}}  & Time                  & 0.75  & 3.49 & 688.69   & 18.76  & 0.83         & 3.45           & 3.71   & $\pmb{0.15}$  & 16.09  & 3.22   & 0.86   & 4.63  & 0.69  & 5.13   \\
                                    & Memory                              & 21.70  & 21.88 & 1976.26  & 22.47  & 21.54        & 21.93          & 22.15  & $\pmb{21.21}$ & 24.18  & 23.71   & 22.89  & 23.77  & 23.24 & 24.61     \\
                                    & Length                              & 14.65 & 12.81 & 16.79  & 12.81  & 15.72        & $\pmb{11.96}$  & $\pmb{11.96}$  & 12.06 & 12.18  & 15.69  & 16.20  & 12.63    & 13.17 & 13.63  \\
                                    & $\mathcal{S}_1 $                    & 32.11 & 40.65 & 109.59  & 48.04  & 128.80        & $\pmb{15.46}$  & $\pmb{15.46}$  & 18.66 & 25.87  & 607.25  & 460.29 & 22.95  & 175.35 & 97.98 \\
                                    & $\mathcal{S}_2 $                    & $\pmb{0.32}$  & 4.50  & 9.25   & 5.32   & 14.67        & 61.66          & 61.66  & 37.50  & 2.86   & 30.41  & 23.05  & 16.12  & 27.38 & 20.88  \\
    \midrule
    \multirow{5}{*}{\makecell{Random \\ Complex}} & Time                  & 0.64  & 2.00   & 348.44    & 32.91  & $\pmb{ 0.12}$      & 0.98           & 1.06   & 0.19  & 25.41  & 1.99   & 0.27   & 3.02  & 0.14    & 6.24   \\
                                    & Memory                              & 21.79 & 21.42 & 1000.14  & 22.77  & 20.88        & 20.96          & 21.18  & $\pmb{20.86}$ & 24.07  & 23.00  & 22.58  & 24.71 & 21.36 & 22.25  \\
                                    & Length                              & 11.84 & 12.19  & 23.59 & 12.19  & 12.19        & $\pmb{11.64}$          & 11.65  & 12.08 & 12.19  & 14.91  & 15.15  & 12.06  & 13.40& 12.71  \\
                                    & $\mathcal{S}_1 $                    & 16.07 & 55.36  & 160.46 & 81.20   & 33.22        & $\pmb{4.30}$            & 10.01  & 26.09 & 55.36  & 605.41 & 286.70 & 16.36  & 260.40 & 91.02 \\
                                    & $\mathcal{S}_2 $                    & $\pmb{0.16}$  & 7.18  &  13.19 & 10.53  & 4.31         & 8.34           & 16.65  & 28.64 & 7.18   & 30.32   & 14.36  & 11.35  & 26.13 & 13.82  \\
    \midrule
    \multirow{5}{*}{\makecell{Simple \\ Maze}}    & Time                  & 1.40   & 99.37 & 621.00  & 124.28 & 1.27         & 108.23         & 135.82 & $\pmb{0.29}$  & 101.11 & 18.37  & 4.74   & 28.47  & 4.18 & 57.02  \\
                                    & Memory                              & 22.57 & 24.43  & 1808.66 & 23.61  & 21.91        & 23.68          & 24.13  & $\pmb{20.72}$ & 26.42  & 25.19     & 24.34     & 25.46  & 23.18 & 24.08   \\
                                    & Length                              & 35.09 & 32.43  & 44.87 & 32.43  & 35.13        & 30.75 & 30.75  & 31.96 & 32.43  & 43.91  & 45.81  & 45.81     & $\pmb{30.62}$ & 34.80  \\
                                    & $\mathcal{S}_1 $                    & 19.38 & 113.77 & 75.66 & 74.92  & 47.39        & $\pmb{15.15}$  & $\pmb{15.15}$  & 23.93 & 29.14  & 581.87 & 368.46 & 28.75 & 246.27 & 97.66 \\
                                    & $\mathcal{S}_2 $                    & $\pmb{0.19}$   & 13.82 & 6.38 & 9.10    & 5.44         & 13.82          & 58.25  & 45.00    & 3.45   & 29.11  & 18.43  & 20.55  & 24.70 & 17.39  \\
    \bottomrule
  \end{tabular}
  \label{performance}
  \vspace{0.5em} 
  \begin{tablenotes}
    \small
    \item Mean: remove the maximum and minimum values and calculate the average of the remaining values.
    \item Time (s): seconds; Memory (MB): megabytes; Length (m): meters; $\mathcal{S}_1$ (deg/m): degrees per meter; $\mathcal{S}_2$ (deg): degrees. 
    \item For all random sampling-based methods, we ran each scenario 500 times and took the average as the final result. 
    \item Simulation environment: Windows 11 OS, python 3.10, AMD Ryzen 5600X CPU, 32GB RAM, RTX 4060Ti.
  \end{tablenotes}
\end{table*}

\begin{figure*}[htp]
  \centering
  \subfigure[Planning process]{
    \includegraphics[width=0.31\linewidth]{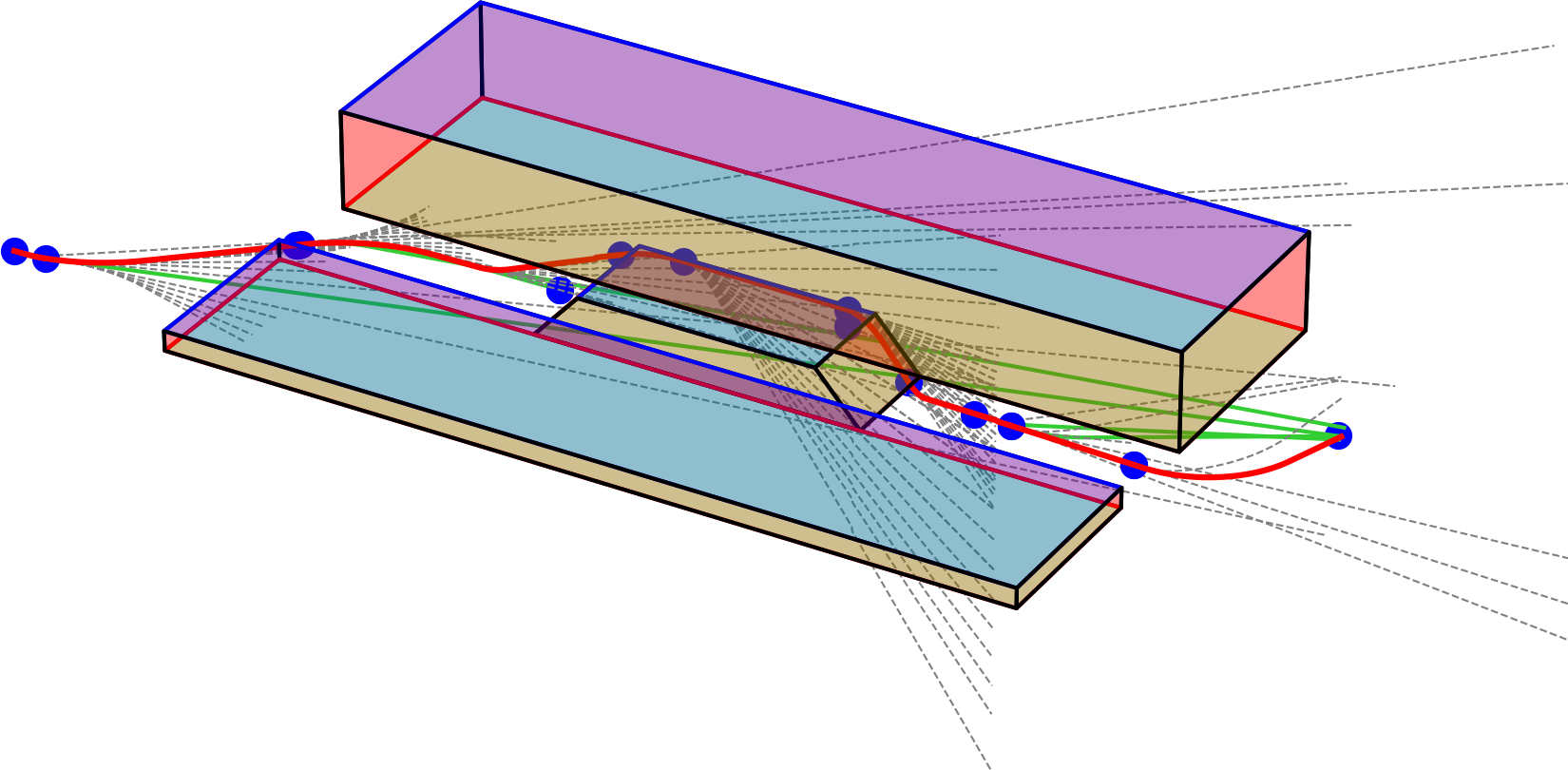}
    \label{fig_3d:a}
  }
  \hspace{-5.5mm} 
  \subfigure[Rviz simulation]{
    \includegraphics[width=0.32\linewidth]{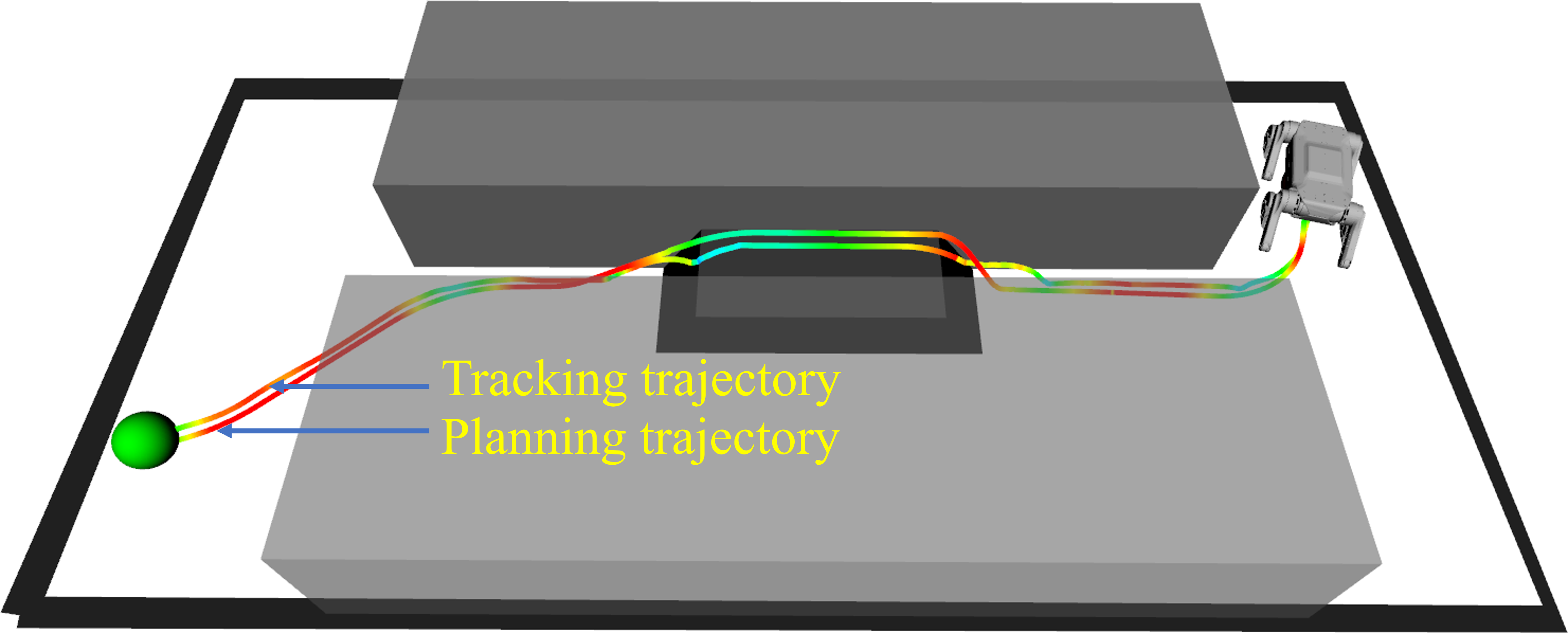}
    \label{fig_3d:b}
  }
  \hspace{-5.5mm} 
  \subfigure[Webots simulation]{
    \includegraphics[width=0.35\linewidth]{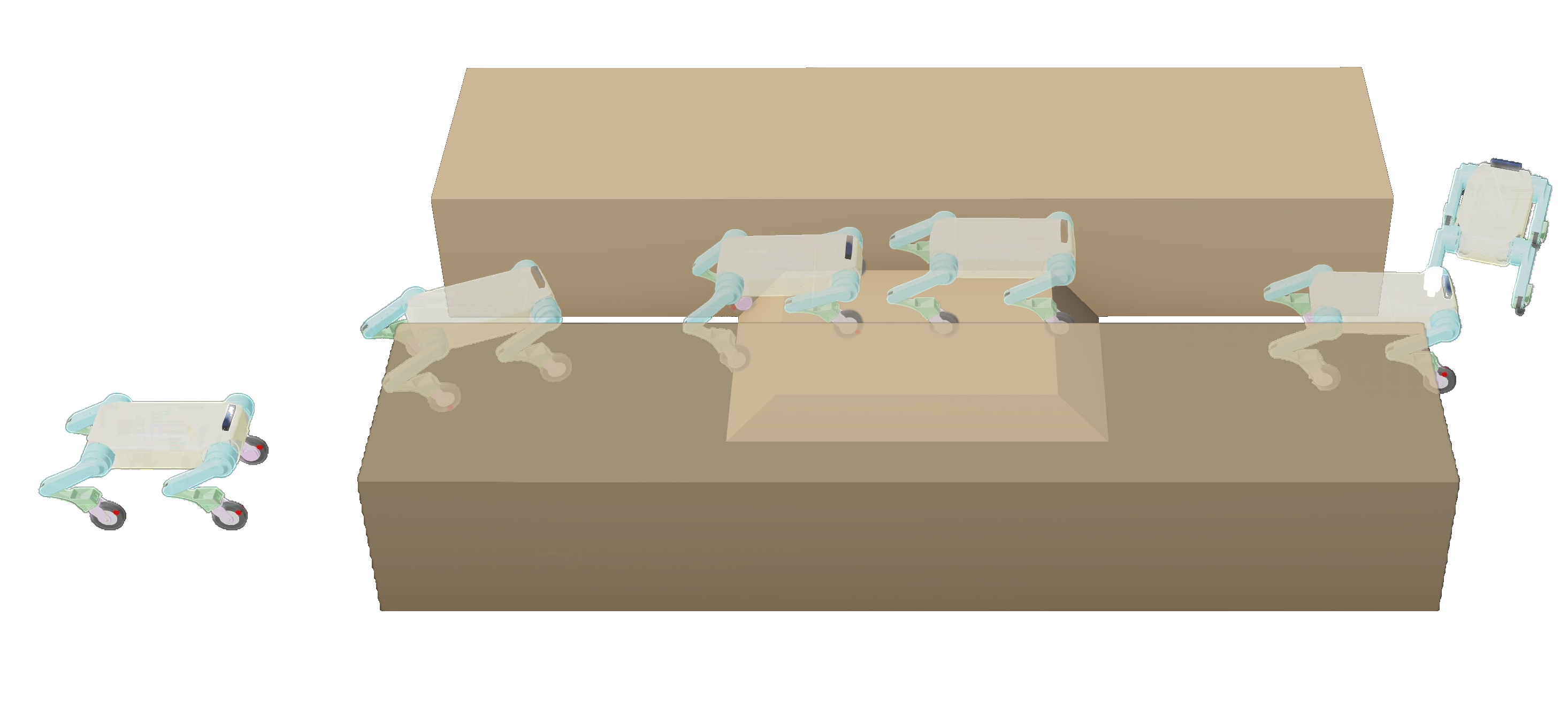}
    \label{fig_3d:c}
  }
  \caption{The demonstration of FDSPC planning in a $2.5$-D terrain-based environment. The planning process is shown in Fig.~\ref{fig_3d:a}, the simulation with velocity and tracking is illustrated in Fig.~\ref{fig_3d:b} (with the tracking trajectory biased vertically to highlight differences), and simulation in Webots is shown in Fig.~\ref{fig_3d:c}. Parameters: $\Delta t = 0.01$, $\rho = 0.4$, $\rho_z = 0.4$, $\theta_{a1} = 0.1$, $l_{add} = 0.5$.}
  \label{fig_3d}
\end{figure*}
\par
However, $\mathcal{S}_1$ may not fully capture the smoothness of the entire path. In cases like Jump Point Search (JPS), the path consists of few discrete waypoints connected by straight lines, the resulting $\mathcal{S}_1$ may indicate high smoothness. Despite this, abrupt changes or sharp turns occur at these waypoints, compromising overall path smoothness. To address this, we introduced a new smoothness metric, $\mathcal{S}_2$, defined as the mean turning angle, which provides a more accurate representation of the path's smoothness.
\par
By exploring feasible trajectories through angle extensions, FDSPC outperforms traditional planners like $A^*$ in solving time on grid maps. Although not the most CPU-efficient, it ranks among the fastest. Unlike sampling-based methods such as RRT, FDSPC maintains stable solving times across different scenarios. In simulations, it requires many integrations to find collision-free paths, but in practice, sensors like LiDAR or depth cameras can provide distance information directly, reducing both memory usage and computation time.
\par
In terms of path length, FDSPC performs averagely in most scenarios but shows a significant increase in the semi-enclosed case due to the added $\theta$ dimension and large orientation deviation, leading to a nearly 180 degrees detour. For $\mathcal{S}_1$ smoothness, FDSPC consistently ranks among the best, slightly outperforming $\theta^*$ and lazy $\theta^*$, and matching JPS. In $\mathcal{S}_2$ smoothness, FDSPC excels significantly thanks to curvature-based planning, outperforming other algorithms by two orders of magnitude.
\par
To verify the feasibility of FDSPC, an obstacle avoidance scenario with uphill and downhill terrain is constructed in a $2.5$-D terrain-based environment, as shown in Fig.~\ref{fig_3d:a}. Simulation results in Rviz and Webots are presented in Fig.~\ref{fig_3d:b} and Fig.~\ref{fig_3d:c}, respectively. The hardware experiment replicates a 3.2m $\times$ 0.3m corridor with a 0.35m-high, 25 degrees ramp in the middle. The obstacle expansion radius is defined as $r_{exp} = \max\{r_{robot} + 0.1,\ r_{robot} \times 1.1\}$, with $r_{robot} = 0.3$m. The wheel-legged robot is equipped with an NUC12 WSK i7, 16-line LiDAR, IMU, RGBD camera, 4 hub motors, 4 steering motors, and 8 joint motors, running ROS Noetic on Ubuntu 20.04. Trajectory tracking is controlled via MPC at 75 Hz and WBC at 200 Hz, as illustrated in Fig.~\ref{fig1}(bottom).
\par
The FDSPC algorithm is most sensitive to the integration step size $\Delta t$ and the linear extension length $l_{add}$, where inappropriate values may lead to planning failures. The angular increment $\theta_{a1}$ primarily affects path smoothness, with larger values significantly degrading curvature continuity. In contrast, the curvature variation rate $\rho$ has relatively minor impact on overall performance. Based on the sensitivity analysis results, the recommended parameter ranges are: $\Delta t = 0.01$-$0.015$~s, $\rho = 0.3$-$0.5$~m$^{-1}$s$^{-1}$, $\theta_{a1} = 0.1$-$0.2$~rad, and $l_{add} = 0.4$-$0.8$~m.

\section{CONCLUSION}
In this letter, we introduced a novel motion planning algorithm FDSPC, based on continuous curvature integration. It explores feasible paths by continuous changes in curvature angles, offering high solution speed, efficient memory usage, shorter path lengths, and exceptional path smoothness. In five typical scenarios, FDSPC demonstrated superior performance, and successfully implemented in obstacle-crossing trajectories on our self-designed wheel-legged robot in a $2.5$-D terrain environment. However, FDSPC has some drawbacks: it's sensitive to parameter settings, may fail to find a path if parameters are unreasonable. In practice, using sensors like LiDAR or depth cameras can reduce integration needs, lowering memory usage and solution time.
\par
In the future, we will improve the FDSPC, reduce adjustable parameters and enhance its completeness, and aim to explore its potential as an effective initial value for mobile robot trajectory optimization.

\bibliographystyle{IEEEtran}

\bibliography{IEEEabrv,ref}

\end{document}